\documentclass{article}

 \usepackage[preprint]{neurips_2026}

\usepackage[utf8]{inputenc} 
\usepackage[T1]{fontenc}    
\usepackage{hyperref}       
\usepackage{url}            
\usepackage{booktabs}       
\usepackage{amsfonts}       
\usepackage{nicefrac}       
\usepackage{microtype}      
\usepackage{xcolor}         
\usepackage{amsmath} 
\usepackage{amssymb}
\usepackage[utf8]{inputenc} 
\usepackage[T1]{fontenc}    
\usepackage{hyperref}       
\usepackage{url}            
\usepackage{booktabs}       
\usepackage{amsfonts}       
\usepackage{nicefrac}       
\usepackage{microtype}      
\usepackage{xcolor}         
\usepackage{algorithm}
\usepackage[table]{xcolor}
\usepackage{colortbl}
\usepackage[normalem]{ulem}
\usepackage{tcolorbox}
\definecolor{thirdcolor}{RGB}{206, 231, 196}
\definecolor{topcolor}{RGB}{252, 236, 196}
\definecolor{secondcolor}{RGB}{223, 235, 253}

\newcommand{\best}[1]{\cellcolor{topcolor}\textbf{#1}}
\newcommand{\second}[1]{\cellcolor{secondcolor}#1}

\newcommand{\bestcap}{\protect\colorbox{topcolor}{\textbf{Best}}}
\newcommand{\secondcap}{\protect\colorbox{secondcolor}{Second-Best}}

\usepackage{enumitem}
\usepackage{algpseudocode}
\usepackage{xcolor}
\usepackage{graphicx}
\usepackage{subcaption}
\usepackage{amsmath,amssymb,amsthm}

\usepackage{booktabs}
\usepackage{multirow}

\usepackage{cancel}
\usepackage{xcolor}
\title{Co-Evolving Skill Generation and Policy Optimization}

\author{%
  Zhiwei Zhang$^{1}$,
  Yudi Lin$^{2}$,
  Nikki Kuang$^{3}$,
  Linlin Wu$^{4}$,
  Xiaomin Li$^{5}$,
  Songtao Liu$^{1}$,
  Fenglong Ma$^{1}$ \\
  $^{1}$The Pennsylvania State University \quad
  $^{2}$Nanyang Technological University \\
  $^{3}$University of California, San Diego \quad
  $^{4}$University of Utah \quad
  $^{5}$Harvard University
}

\begin{document}

\maketitle

\begin{abstract}

Skill-augmented reinforcement learning improves language agents by storing reusable procedural knowledge acquired from past experience. Existing methods typically use strong language models to analyze trajectories, generate skills, and update a retrievable skill bank during online training. However, they rarely assess whether a newly generated skill is useful before it is stored and reused. We find that this assumption is unreliable: even skills generated by proprietary frontier LLMs exhibit highly mixed utility, with many providing little benefit or even degrading performance. Once such skills enter the bank, their effects are difficult to identify, because subsequent rollout feedback is delayed and usually reflects the combined effect of multiple retrieved skills rather than the marginal contribution of any individual skill.
We propose an online reinforcement learning framework for pre-storage skill validation. The framework estimates whether a candidate skill contributes useful information beyond the skills already retrieved for the current task. It uses the standard rollout budget to form two matched groups under the same task and retrieval context: base rollouts conditioned on the currently retrieved skills, and skill-augmented rollouts conditioned on the same skills plus one candidate skill induced from the base trajectories. The reward gap between these two groups estimates the candidate skill’s context-dependent marginal utility, enabling the framework to promote useful skills while filtering ineffective or harmful ones without additional rollout overhead.
The framework further uses this marginal-utility signal to train the policy itself as a skill generator, reducing reliance on repeated calls to proprietary models. The learned skill-generation likelihood serves as a context-dependent score for retrieval-time reranking and outdated-skill pruning as the policy evolves. Experiments on interactive decision-making and search-augmented question answering benchmarks show that the proposed framework outperforms prior skill-augmented RL methods while improving skill quality and avoiding costly proprietary skill-generation APIs.
Ours code is available at \url{https://github.com/zzwjames/skill_augmented_agent}.

\end{abstract}

\section{Introduction}
Agentic large language models (LLMs)~\citep{zhou2026externalization,zhou2026memento,yu2026agemem,xu2026ael,buyya2026agentic,zhang2025landscape,plaat2025agentic} are increasingly applied to complex tasks
reasoning~\citep{wei2026agentic,chen2026think,hao2026brain,feng2026idrbench,li2026benchmark,wu2025agentic,zhao2025llm}, tool use~\citep{xu2026agentskills,jiang2026sok,wang2026skillx,xia2026skillrl,yang2026tooltree,hu2026agentic,lin2026w}, or interaction with external
environments. To improve agents in such settings, recent work has introduced
\emph{skills}: reusable procedural knowledge that equips LLMs with
domain-specific capabilities and guides their behavior across related tasks
~\citep{jiang2026sok,xu2026agentskills, zhang2026equipping}. 
Skills typically encode reusable procedures, such as action patterns or decision
rules, and can be represented in different forms, including natural-language
guidance, executable programs, or memory units
~\citep{zhang2026equipping,huang2025cascade,skillweaver,yang2025exif,zhou2025pae,fang2025memp,cao2025reme}.
They may be manually designed or automatically induced from agent experience,
and can be retrieved to condition future decisions on useful procedural
knowledge.

Recent studies have explored different ways to use skills for agent improvement.
One line of work keeps the base model fixed and improves external skill memory
or task-specific context. For example, MCE~\citep{ye2026meta} uses a stronger
model to refine skills from trajectory-level feedback. 
Another line of work studies skill evolution
together with policy optimization. SkillRL~\citep{xia2026skillrl} updates a
retrievable skill bank during online RL. Skill0~\citep{lu2026skill0} focuses on
skill internalization, progressively transferring skill knowledge into the
policy so that the agent can rely less on runtime skill retrieval. D2Skill
~\citep{tu2026dynamic} maintains skills at multiple granularities and updates
them with utility signals from rollout outcomes.

Despite this progress, we identify a critical issue in existing skill-augmented
RL methods: \textit{Newly generated skills are often inserted into the skill bank before
their \textbf{usefulness} is explicitly validated.} In particular, many existing methods
\citep{xia2026skillrl,lu2026skill0,tu2026dynamic} rely on proprietary frontier
LLMs to analyze trajectories and generate skills, implicitly assuming that
skills produced by stronger models are reliable. However, our preliminary
experiments in Sec.~\ref{sec:pre} show that even skills generated by GPT-5.4 exhibit highly mixed utility. While a small subset improves performance, many provide little benefit or even mislead the agent, and the average utility stays close to zero throughout training. This is especially concerning because costly frontier-LLM API calls still produce many skills that fail to help or even interfere with policy learning.
Although some prior methods \citep{tu2026dynamic, zhou2026memento} attempt to track skill utility, their feedback is delayed. For example, D2Skill~\citep{tu2026dynamic} estimates skill usefulness from later rollouts after a skill has already been stored and retrieved. By the time a low-quality skill is identified, it has already misled the agent and slowed down policy learning.

Furthermore, skill usefulness changes as the policy evolves. As the policy
becomes stronger during online RL, some previously useful skills may become
outdated or redundant. Keeping such skills in the bank can introduce retrieval
noise, because they may occupy slots that could otherwise be assigned to more
useful skills. Effective skill maintenance therefore requires identifying skills
whose value has decreased over training.
A natural way to measure the current value of a skill is to compare rollouts with
and without that skill, but doing so naively would require additional
counterfactual rollouts. Prior work on memory or skill valuation
\citep{tu2026dynamic, zhou2026memento} has explored learned scoring functions or
delayed utility estimates, but these signals often fail to isolate the marginal
contribution of an individual skill. In retrieval-based skill augmentation,
skills are usually used jointly, so the observed performance reflects the
aggregate effect of multiple co-retrieved skills rather than the value of any
single skill.
This matters because an ideal skill-utility measure should be context-dependent. A skill that
is useful in isolation may become redundant when similar skills are already
retrieved, while another skill may be valuable because it complements the current
retrieval context. Thus, both skill maintenance and retrieval-time selection
require estimating whether a skill provides marginal benefit beyond the skills
already in the bank or already selected for the current query.

{These observations reveal three key challenges for skill-augmented online RL}:
\textbf{(1)} how to estimate the marginal utility of newly generated skills
without incurring additional rollout cost; \textbf{(2)} how to score a skill by
its marginal contribution beyond the currently retrieved skill context, rather
than by its standalone relevance; and \textbf{(3)} how to reduce reliance on
costly proprietary LLM APIs for skill generation without sacrificing skill
quality. 
To tackle these challenges, we propose \underline{\textbf{S}}kill-\underline{\textbf{A}}ugmented \underline{\textbf{P}}olicy \underline{\textbf{O}}ptimization
(SAPO), a novel online RL framework that validates skills before storage and
uses their context-dependent utility to improve skill generation, maintenance,
and retrieval. 

SAPO first splits the standard rollout budget into two matched
groups for prospective skill validation. During training, SAPO retrieves existing
skills for a task and uses the first part of the rollout budget to generate base
rollouts. These rollouts show how the current policy behaves under the existing
skill context and provide evidence for inducing a candidate skill. SAPO then uses
the remaining rollout budget to generate skill-augmented rollouts under the same
task and retrieved skill context, with only the candidate skill added. Because
the two rollout groups share the same task and retrieved skill context and differ
only in the presence of the candidate skill, their reward gap estimates the
candidate's marginal contribution beyond the retrieved skills, without
requiring additional rollouts. SAPO uses this context-dependent marginal utility
to curate the skill bank before storage: useful and non-redundant skills are
promoted into long-term memory, while low-utility skills are discarded before
they can affect future retrieval and policy learning.
The same utility signal further trains the policy itself as a stronger skill generator. 
This avoids repeated dependence on proprietary frontier LLMs for skill generation. 
Once trained with utility feedback, the policy's probability of generating a skill serves as a quality score. SAPO uses this context-dependent score for long-term maintenance and retrieval-time selection, pruning outdated skills and reranking candidate skills during retrieval.

We evaluate SAPO across diverse interactive decision-making and
search-augmented question answering tasks. SAPO consistently outperforms prior
skill-augmented RL methods while avoiding costly API calls to proprietary
models. 
Further analyses show improved training dynamics, increasing skill
utility, and the effectiveness of SAPO's major components.
Our main contributions are: 
\textbf{(1)} we reveal the critical issue that prior methods overlook the mixed utility of generated skills, allowing low-quality skills to mislead agent learning;
\textbf{(2)} we propose SAPO, a novel online RL framework that provides skill utility estimation without additional overhead and trains the policy as both an agent and a skill generator; 
and \textbf{(3)} extensive experiments show that SAPO outperforms existing methods while avoiding costly API calls to proprietary models. 

\section{Related Work}

\paragraph{Agent Skills.}
In autonomous agents, skills are reusable procedural knowledge for temporally
extended, goal-directed behaviors beyond one-step action generation
~\citep{jiang2026sok, xu2026agentskills}. They encode action patterns, decision
procedures, or tool-use workflows across related tasks
~\citep{zhang2026equipping, wang_voyager_2024}, and may appear as natural-language
instructions~\citep{liu2024skillact}, executable programs~\citep{wang2025asi,
huang2025cascade}, APIs~\citep{skillweaver}, trajectories~\citep{yang2025exif,
zhou2025pae}, or memory units~\citep{fang2025memp, cao2025reme}. These forms
connect high-level task intents to low-level actions, making skills an
operational form of procedural memory~\citep{wu2026survey,
zhou2026externalization, zhang2026experiencecompression}.
Prior work obtains skills through manual design~\citep{zhang2026equipping},
demonstrations~\citep{liu2024skillact}, repository mining
~\citep{bi2026skillmining}, knowledge-base construction
~\citep{wang2026skillx, shen2026skillfoundry}, exploration~\citep{yang2026autoskill}, or trajectory/feedback distillation
~\citep{ni2026trace2skill, xia2026skillrl}. Recent efforts further
extend skill repositories through external policy guidance~\citep{
wang2025reinforcement, li2026arise}, parameter internalization
~\citep{lu2026skill0}, co-evolutionary verification
~\citep{zhang2026coevoskills}, collective evolution~\citep{ma2026skillclaw}, and
lifelong self-evolution~\citep{yang2026autoskill}. As repositories grow, the challenge shifts from skill generation to skill
valuation~\citep{
li2026skillsbench, zhong2026skilllearnbench, liu2026skillswild,
wang2026skilltester, zhang2026skillflow}.

\paragraph{Agent Memory.}

Agent memory stores past experience in editable and retrievable forms for future
use~\citep{suttonWelcomeEraExperience2025,
du2026memorysurvey}. Prior work studies memory at different levels:
\emph{case-based memory} stores raw trajectories, solutions, or examples for
similar future tasks~\citep{
zhou2508memento, chen2025scalingagentlearningexperience,
agent-early-experience, zhang2026memrl, fang2026trajectorymemory};
\emph{strategy-based memory} summarizes interactions into reusable insights,
workflows, or reasoning patterns~\citep{ouyang2025reasoningbankscalingagentselfevolving,
huang_r2d2_2025,
suzgun2025dynamiccheatsheettesttimelearning,
cai2025flexcontinuousagentevolution, xu2026ael}; and
\emph{skill-based memory} stores callable knowledge such as code, functions, and
APIs that map high-level plans to concrete actions~\citep{zhang_darwin_2025, skillweaver, fang2025memp, wang2025asi, han_legomem_2025, yang2026autoskill,
ni2026trace2skill}. 
Beyond storing experience, recent work studies how memory should be organized
and updated over time, including unified memory pipelines~\citep{
tang2025agentkbleveragingcrossdomain, huang_r2d2_2025, Zhang2025GMemory,
wu2025evolverselfevolvingllmagents, yu2026agemem}, graph-structured
memory~\citep{yang2026graphmemory}, and memory management or evolution through
consolidation, updating, forgetting, and RL-based construction~\citep{
zhang2025memevolvemetaevolutionagentmemory,
zhai2025agentevolverefficientselfevolvingagent, cai2025experiencedriven,
yan2025memory, khanda2026crystallization}.

\section{Preliminary}
\label{sec:pre}

In this section, we present two preliminary findings that motivate our method design.

\textbf{Experimental Setup.}
We follow the experimental setting of SkillRL~\citep{xia2026skillrl} and report results on \textsc{ALFWorld}\citep{shridhar2020alfworld} and \textsc{WebShop}\citep{yao2022webshop}. The base model used in all experiments is \texttt{Qwen2.5-7B-Instruct}~\citep{yang2025qwen3}.
\begin{figure}[h]
    \centering
    \begin{subfigure}[t]{0.495\textwidth}
        \centering
        \includegraphics[width=\textwidth]{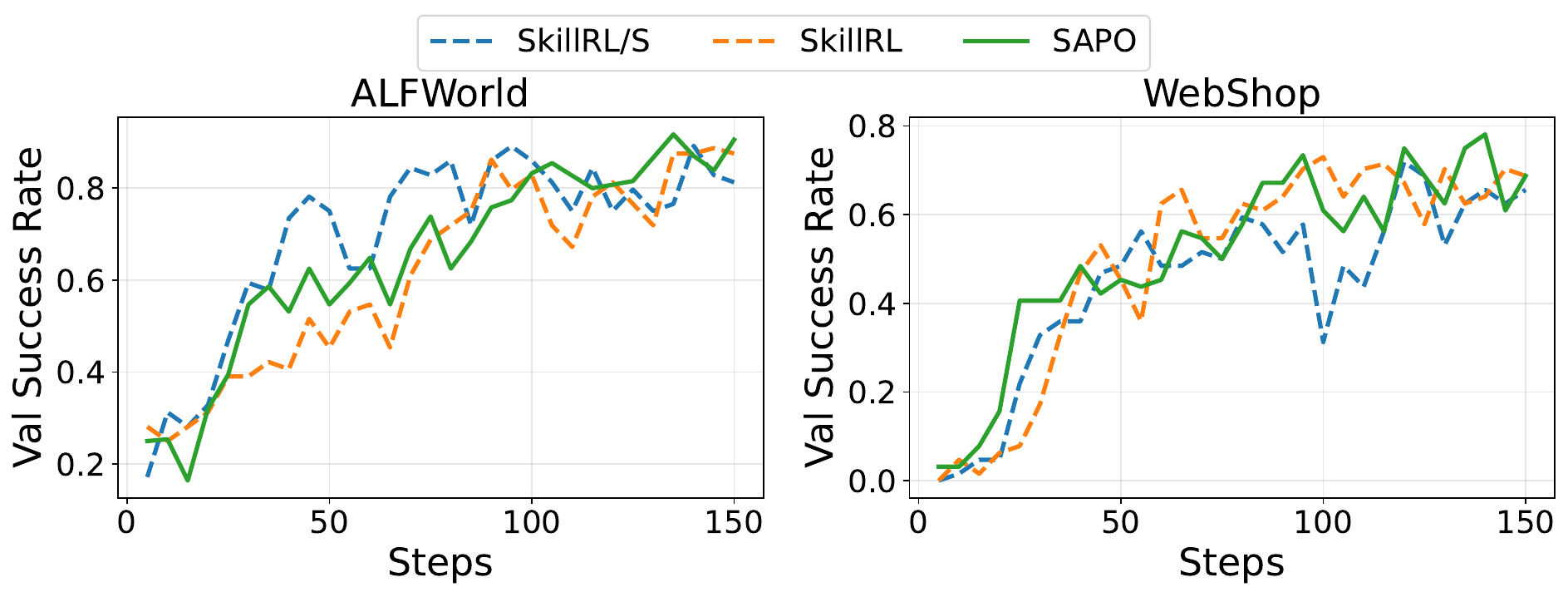}
        \label{fig:alf_web_success}
    \end{subfigure}
    \hfill
    \begin{subfigure}[t]{0.495\textwidth}
        \centering
        \includegraphics[width=\textwidth]{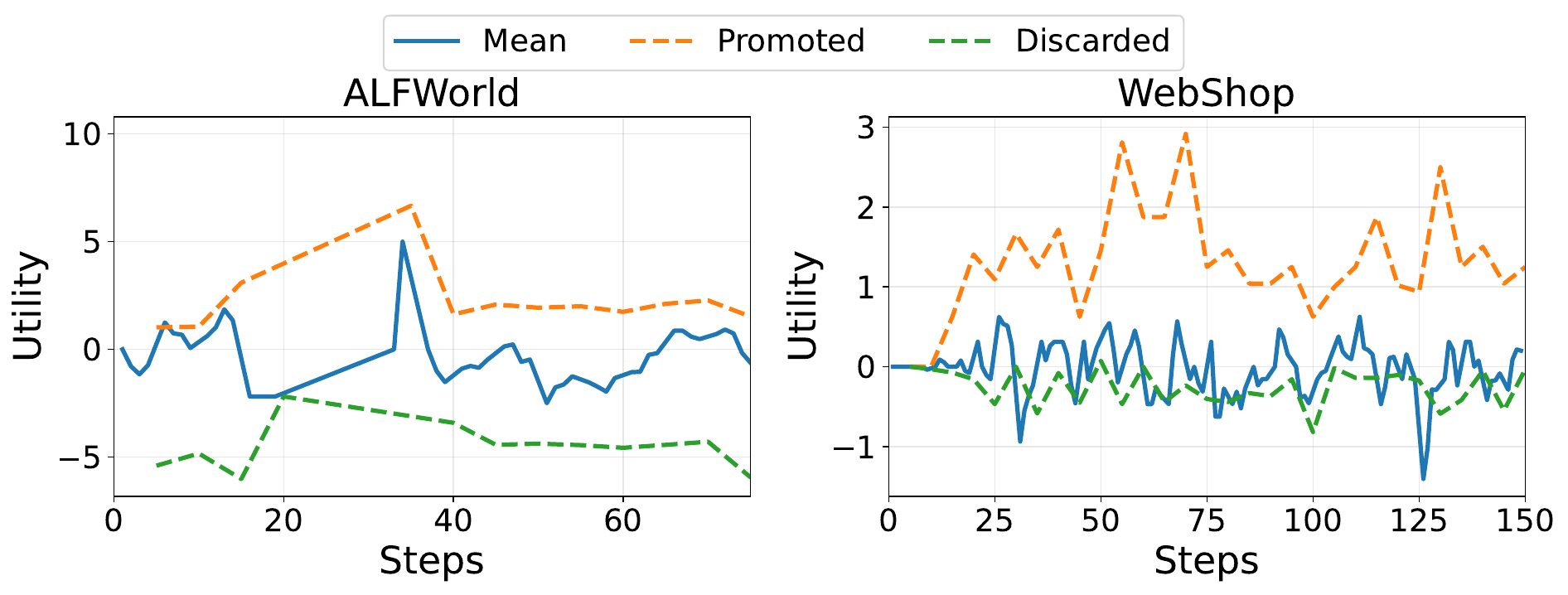}
        \label{fig:alf_web_utility}
    \end{subfigure}
    \caption{Training dynamics on ALFWorld and WebShop. Left: validation success rate. Right: utility of GPT-generated skills.}
    \label{fig:alf_web_main}
\end{figure}
\paragraph{Skill evolution brings only marginal improvement over a no-evolution variant.}
We compare SkillRL with SkillRL/S, a no-evolution variant with a fixed skill bank. As shown in Fig.~\ref{fig:alf_web_main}, SkillRL provides only limited gains on both \textsc{ALFWorld} and \textsc{WebShop}, and even reaches its best \textsc{ALFWorld} performance more slowly than SkillRL/S. This suggests that naively evolving the skill bank does not necessarily improve learning efficiency.

\paragraph{Even GPT-generated skills are mixed, with near-zero average utility.}
To understand this limited gain, we estimate the marginal utility of GPT-generated skills during training. For each prompt, we sample $G$ base rollouts, ask GPT-5.4 to induce a skill from the observed failures, and then sample another $G$ rollouts with the generated skill added. The reward gap between the two groups is used as the skill's marginal utility, which we average across prompts at each training step.
Every 5 training steps, we further split generated skills into \emph{promoted} skills, defined as positive-utility skills in the top 20\%, and \emph{discarded} skills, containing the rest.

As shown in Fig.~\ref{fig:alf_web_main}, the mean marginal utility of GPT-generated skills stays close to zero on both benchmarks, despite a clear gap between promoted and discarded skills. This indicates that generated skills are highly mixed: a small subset is useful, while many are ineffective or harmful. The limited improvement of SkillRL may therefore stem from admitting low-utility skills into the bank, where they can mislead retrieval and slow learning. 
Appendix~\ref{app:claude_utility} shows a similar pattern with Claude-Opus-4.6 as the skill generator.

Based on the preliminary findings above and our analysis of existing work, we identify three limitations of
existing skill-augmented agents in online reinforcement learning settings:
\begin{itemize}[leftmargin=*]
    \item \textbf{Direct Injection of Unvalidated Skills.} Existing methods often insert generated skills into the bank without validation, allowing low-quality or harmful skills to be stored and later retrieved. Although some methods~\citep{zhou2026memento, tu2026dynamic} track skill utility after storage, this feedback is delayed: a low-quality skill has already misled the agent and harmed policy learning before being identified.
    \item \textbf{Lack of Context-Dependent Attribution for Individual Skills.}
Ideally, skill utility should be context-dependent: a skill may be redundant when similar
skills are retrieved, or useful when it complements the current context.
However, existing methods either estimate skill utility~\citep{tu2026dynamic} or
train utility functions~\citep{zhou2026memento, zhang2026memrl} from
set-level signals, where the observed return reflects all retrieved skills
jointly. Thus, they cannot determine an individual skill's marginal value in
that context.
    \item \textbf{High API Cost of Skill Generation.}
Existing methods~\citep{xia2026skillrl, tu2026dynamic, lu2026skill0} rely on
proprietary frontier LLMs to generate skills during training, incurring high
costs across repeated online RL updates, while many skills have negative utility. Appendix~\ref{app:api_cost} analyzes API costs of existing methods.
\end{itemize}

\section{Skill-Augmented Policy Optimization}
\label{sec:sapo}

\begin{figure}[t]
    \centering
    \includegraphics[width=0.8\linewidth]{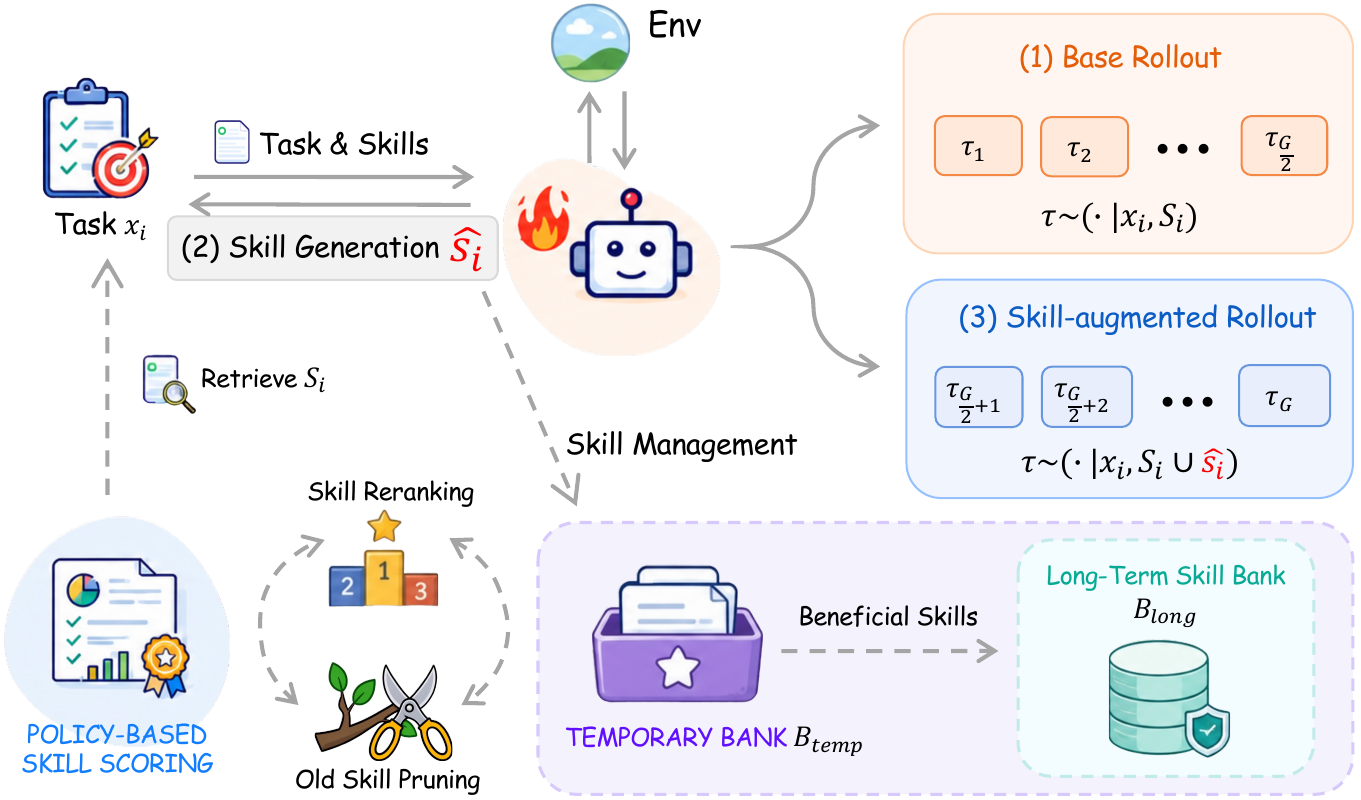}
    \caption{Framework of SAPO. SAPO first obtains base rollouts $\mathcal{Y}_{i,j}^{\mathrm{base}}$, then generates a candidate skill $\hat{s}_i$, and validates it with skill-augmented rollouts $\mathcal{Y}_{i,j}^{\mathrm{skill}}$.}
    \label{fig:frame}
\end{figure}

As shown in Figure~\ref{fig:frame}, we propose Skill-Augmented Policy Optimization (SAPO), an online
reinforcement learning framework that jointly performs \emph{skill induction},
\emph{skill validation}, and \emph{skill maintenance}. SAPO optimizes the agent policy with rewards from collected rollouts while
estimating the marginal utility of newly induced skills online. This utility
signal curates the skill bank and trains the same policy as a stronger skill
generator, whose skill-generation likelihood is further reused as a skill score for maintenance and retrieval-time reranking.
Next, we describe each component in detail.

\subsection{Online Rollouts and Skill Induction}
SAPO operates on a \emph{skill induction unit} $\mathcal{X}_i=\{x_{i,j}\}_{j=1}^K$, which contains either a single query ($K=1$) or a small group of semantically similar queries ($K>1$). Each induction unit produces one candidate skill $\hat{s}_i$.
SAPO maintains two skill banks: a long-term bank $\mathcal{B}_{\mathrm{long}}$ for validated skills and a temporary bank $\mathcal{B}_{\mathrm{temp}}$ for candidates awaiting promotion.
For each query $x_{i,j} \in \mathcal{X}_i$, SAPO first retrieves a set of relevant existing skills,
$\mathcal{S}_{i,j} \leftarrow \mathrm{Retr}(x_{i,j}, \mathcal{B}_{\mathrm{long}} \cup \mathcal{B}_{\mathrm{temp}})$,
and allocates a rollout budget of $G$ trajectories, split into two matched halves. The first half consists of \emph{base rollouts} generated under the retrieved skill context alone:
\begin{equation}
\mathcal{Y}_{i,j}^{\mathrm{base}} \sim \pi_\theta(\cdot \mid x_{i,j}, \mathcal{S}_{i,j}).
\label{eq:base_traj_group}
\end{equation}

Based on the induction evidence
$
\{(x_{i,j}, \mathcal{S}_{i,j}, \mathcal{Y}_{i,j}^{\mathrm{base}})\}_{j=1}^K,
$
the policy then induces a candidate skill
\begin{equation}
\hat{s}_i \sim 
\pi_\theta\!\left(
\cdot \mid 
\mathcal{X}_i,\;
\{\mathcal{S}_{i,j}\}_{j=1}^K, X_{\mathrm{skill}},\;
\{\mathcal{Y}_{i,j}^{\mathrm{base}}\}_{j=1}^K\;
\right),
\label{eq:shared_skill_gen}
\end{equation}
where $X_{\mathrm{skill}}$ is a skill-generation prompt template (Appendix~\ref{app:prompt}). Intuitively, the base trajectories reveal both the policy's current capabilities under the retrieved context and the remaining failure patterns, enabling $\hat{s}_i$ to encode reusable guidance for future rollouts. 
However, as shown in Sec.~\ref{sec:pre}, even GPT-generated skills are mixed in quality. SAPO therefore validates each candidate skill before storage by using the remaining rollout budget to generate \emph{skill-augmented rollouts}:
\begin{equation}
\mathcal{Y}_{i,j}^{\mathrm{skill}} \sim 
\pi_\theta(\cdot \mid x_{i,j}, \mathcal{S}_{i,j} \cup \{\hat{s}_i\}),
\label{eq:skill_aug_group}
\end{equation}
so that $(\mathcal{Y}_{i,j}^{\mathrm{base}}, \mathcal{Y}_{i,j}^{\mathrm{skill}})$ forms a matched rollout pair under the same query and retrieved context, differing only in whether the induced skill is included. The collected base and skill-augmented rollouts are used to optimize the agent policy with GRPO \citep{shao2024deepseekmath}.

\subsection{Skill Utility Estimation and Bank Update} 

Given the base rollouts
$\mathcal{Y}_{i,j}^{\mathrm{base}}$
and the skill-augmented rollouts
$\mathcal{Y}_{i,j}^{\mathrm{skill}}$,
SAPO estimates the marginal utility of the candidate skill $\hat{s}_i$ by
comparing the rewards of the two rollout sets.

\paragraph{Intra-prompt Utility.}
For a candidate skill $\hat{s}_i$ and a source query $x_{i,j}$, we define its
prompt-specific marginal utility as
\begin{equation}
u(x_{i,j}, \hat{s}_i)
=
\frac{1}{|\mathcal{Y}_{i,j}^{\mathrm{skill}}|}
\sum_{y \in \mathcal{Y}_{i,j}^{\mathrm{skill}}} r(x_{i,j}, y)
-
\frac{1}{|\mathcal{Y}_{i,j}^{\mathrm{base}}|}
\sum_{y \in \mathcal{Y}_{i,j}^{\mathrm{base}}} r(x_{i,j}, y),
\label{eq:instance_utility}
\end{equation}
where $r(x,y)$ denotes the task reward. This quantity estimates the conditional marginal effect of adding $\hat{s}_i$ under the query $x_{i,j}$ and the retrieved skill context $\mathcal{S}_{i,j}$.

\paragraph{Cross-prompt Utility.}
When $K>1$, the candidate skill $\hat{s}_i$ is induced from an induction unit
$\mathcal{X}_i=\{x_{i,j}\}_{j=1}^{K}$. We therefore aggregate its utilities over
the queries in the same unit:
\begin{equation}
U_{\hat{s}_i}
=
\frac{1}{K}
\sum_{j=1}^{K}
u(x_{i,j}, \hat{s}_i).
\label{eq:group_utility}
\end{equation}
This cross-prompt utility measures whether the shared skill generalizes across related queries, and reduces to
$U_{\hat{s}_i}=u(x_{i,1},\hat{s}_i)$ when $K=1$. SAPO then stores
$(\hat{s}_i,U_{\hat{s}_i})$ in $\mathcal{B}_{\mathrm{temp}}$ for later promotion.

\paragraph{Skill Promotion and Bank Update.}
At the end of each temporary horizon, SAPO promotes a temporary skill to the
long-term bank only if it has positive utility, ranks among the top 
fraction of temporary skills, and is sufficiently distinct from existing
long-term skills:
\begin{equation}
U_s > 0,
\quad
s \in \mathrm{Top}_{\rho}\!\left(\mathcal{B}_{\mathrm{temp}}; U\right),
\quad
\max_{s' \in \mathcal{B}_{\mathrm{long}}}
\mathrm{Sim}(s,s') < \gamma .
\label{eq:skill_promotion}
\end{equation}
Here, $\mathrm{Top}_{\rho}(\mathcal{B}_{\mathrm{temp}}; U)$ selects the top
$\rho$ fraction by validation utility, and $\gamma$ controls the novelty
threshold against the long-term bank. Promoted skills are
added to $\mathcal{B}_{\mathrm{long}}$, while the remaining temporary skills are
discarded before the next horizon.

\subsection{Policy as a Skill Generator and Scorer}
\label{sec:policy_generator_scorer}

The skill utility in Eq.~\eqref{eq:instance_utility} provides direct supervision
for each newly induced skill without extra rollout budget, and measures the
skill's marginal contribution beyond the existing retrieved skills. However,
using this utility only to filter or track low-quality skills leaves two issues
unresolved. First, existing methods often rely on costly closed-source LLMs for
skill generation. Second, evaluating stored or retrieved skills for outdated-skill
pruning and retrieval-time selection through additional counterfactual rollouts
would be computationally expensive.
SAPO therefore reuses the utility signal to train the policy itself as both
a skill generator and a skill scorer.

\paragraph{Skill Generator Training.}
The utilities in Eq.~\eqref{eq:instance_utility} provide direct supervision for
the skill generator. Inspired by W-REINFORCE~\citep{zhusurprising}, SAPO trains
the generator with an asymmetric utility-weighted objective: positive-utility
skills are reinforced, while negative-utility skills are suppressed more
strongly.
Specifically, let
$\ell_\theta(\hat{s}_i;x_i)=\log \pi_\theta(\hat{s}_i\mid x_i,\mathcal{S}_i,
X_{\mathrm{skill}},\mathcal{Y}_i^{\mathrm{base}})$ denote the log-probability of
generating $\hat{s}_i$ from the retrieved skills and base rollouts of $x_i$.
SAPO optimizes
\begin{equation}
\mathcal{L}_{\mathrm{gen}}(\theta)
=
-
\mathbb{E}_{x_i}
\left[
\lambda [u(x_i,\hat{s}_i)]_+\,
\ell_\theta(\hat{s}_i;x_i)
\right]
+
\mathbb{E}_{x_i}
\left[
[-u(x_i,\hat{s}_i)]_+\,
\ell_\theta(\hat{s}_i;x_i)
\right],
\label{eq:utility_weighted_generator}
\end{equation}
where $[a]_+=\max(a,0)$ and $\lambda\in[0,1]$ down-weights positive
reinforcement.
We use the intra-prompt utility rather than a cross-prompt utility to supervise
the generator. First, even when $K>1$ and the generator conditions on multiple
queries and their corresponding rollouts, the skill-generation interface remains
the same as in the single-query case: the grouped setting only provides more task
descriptions and base rollouts, while the prompt structure and output format are
unchanged. More details are provided in Appendix~\ref{app:prompt}.
Thus, intra-prompt utility teaches the generator to extract a useful skill from
the evidence available in the current prompt, and this ability naturally extends
to grouped-query skill induction. The effectiveness of this design is further
supported by Sec.~\ref{sec:skill_utility}.
Second, skill maintenance and retrieval-time
reranking are query-specific decisions. A cross-prompt utility can blur this
query-specific value: a skill that is unhelpful for the current query may still
receive a high score because it is useful for other queries. Intra-prompt utility
therefore provides a more faithful supervision signal for both generating and
selecting skills.

\paragraph{Policy-Based Skill Scoring.}
Eq.~\eqref{eq:utility_weighted_generator} trains the policy to assign higher
likelihood to positive-utility skills and lower likelihood to negative-utility
skills. Thus, the skill-generation likelihood itself can serve as a learned
skill-usefulness score. Given a task $x$, skill context $\mathcal{S}$, optional
rollout evidence $\mathcal{Y}$, and candidate skill $s$, SAPO defines
\begin{equation}
\texttt{Score}_{\pi_\theta}
\left(
x,\mathcal{S},X_{\mathrm{skill}},[\mathcal{Y}],s
\right)
=
\frac{1}{|s|}
\sum_{l=1}^{|s|}
\log \pi_\theta\!\left(
s^{(l)}
\mid
x,
\mathcal{S},
X_{\mathrm{skill}},
[\mathcal{Y}],
s^{(<l)}
\right),
\label{eq:policy_score}
\end{equation}
where $[\mathcal{Y}]$ denotes an optional rollout input. During training, the
full-mode score is conditioned on the collected base rollouts
$\mathcal{Y}^{\mathrm{base}}_i$. At maintenance or retrieval time, however,
collecting rollouts for every candidate skill would be costly, so SAPO omits
$[\mathcal{Y}]$. To transfer the preferences learned in the full mode to this
efficient reduced mode, SAPO distills the full distribution into the
reduced-input distribution:
\begin{equation}
\mathcal{L}_{\mathrm{KD}}(\theta)
=
\mathbb{E}_{x_i}
\left[
\mathrm{KL}\left(
\mathrm{sg}\!\left[
\pi_\theta(\cdot \mid x_i,\mathcal{S}_i,X_{\mathrm{skill}},
\mathcal{Y}^{\mathrm{base}}_i)
\right]
\;\middle\|\;
\pi_\theta(\cdot \mid x_i,\mathcal{S}_i,X_{\mathrm{skill}})
\right)
\right],
\label{eq:kd}
\end{equation}
where $\mathrm{sg}[\cdot]$ denotes stop-gradient.

\paragraph{Long-Term Skill Maintenance.}
When the long-term skill bank reaches its capacity limit, SAPO removes outdated
or low-value skills using the reduced-input skill-likelihood score in
Eq.~\eqref{eq:policy_score}. For each old skill $s$, SAPO constructs a set of
relevant evaluation prompts $\mathcal{P}(s)$. For each prompt
$x\in\mathcal{P}(s)$, SAPO retrieves a contextual skill sequence
$\mathbf{S}^{\mathrm{ctx}}(x,s)$ from the current bank while excluding $s$, and
scores the old skill with
$\texttt{Score}_{\pi_\theta}
(x,\mathbf{S}^{\mathrm{ctx}}(x,s),X_{\mathrm{skill}},s)$.
The scores are averaged over $\mathcal{P}(s)$, and skills with low average
scores are removed from the long-term bank.

\paragraph{Retrieval-Time Skill Selection.}
Whenever skill retrieval is needed, SAPO first retrieves a candidate pool
$\mathcal{C}(x)$ using similarity-based retrieval and then reranks it with the
reduced-input skill-likelihood score in Eq.~\eqref{eq:policy_score}. Skills are
selected sequentially: at step $k$, given already selected skills
$\mathbf{S}_{k-1}=(s_1,\ldots,s_{k-1})$ with $\mathbf{S}_0=\emptyset$, SAPO
scores each remaining candidate
$s\in\mathcal{C}(x)\setminus\{s_1,\ldots,s_{k-1}\}$ using
$\texttt{Score}_{\pi_\theta}(x,\mathbf{S}_{k-1},X_{\mathrm{skill}},s)$ and
selects the highest-scoring one as $s_k$. The process repeats until the
retrieval budget is reached. The full SAPO algorithm is provided in
Appendix~\ref{app:algo}.

\section{Experiments}
\label{sec:exp}

\subsection{Experimental Setup}

\noindent\textbf{Benchmarks.}
We evaluate SAPO on three benchmark families: \textsc{ALFWorld}
~\citep{shridhar2020alfworld}, \textsc{WebShop}~\citep{yao2022webshop}, and
search-augmented question answering. \textsc{ALFWorld} tests embodied household
tasks through text-based interaction, while \textsc{WebShop} evaluates web
navigation for product search and purchase under user-specified constraints. For search-augmented QA, we consider both single-hop datasets, including NQ \citep{kwiatkowski2019natural}, TriviaQA \citep{joshi2017triviaqa}, and PopQA \citep{mallen2023not}, and multi-hop datasets, including HotpotQA \citep{yang2018hotpotqa}, 2Wiki \citep{ho2020constructing}, MuSiQue \citep{trivedi2022musique}, and Bamboogle \citep{press2023measuring}.

\noindent\textbf{Baselines.}
We compare SAPO against five categories of baselines. First, we include strong closed-source LLMs, including GPT-4o and Gemini-2.5-Pro \citep{comanici2025gemini}, which represent state-of-the-art general-purpose reasoning models. Second, we consider prompt-based and memory-based agentic methods, including ReAct \citep{yao2022react}, Reflexion \citep{shinn2023reflexion}, Mem0 \citep{chhikara2025mem0}, ExpeL \citep{zhao2024expel}, and MemP \citep{fang2025memp}. Third, we include general RL baselines, including RLOO \citep{ahmadian2024back} and GRPO \citep{shao2024deepseekmath}. Fourth, we compare against memory-augmented RL methods, including EvolveR \citep{wu2025evolverselfevolvingllmagents}, MemRL \citep{zhang2026memrl}, and SimpleMem \citep{liu2026simplemem}. Fifth, we compare with prior skill-augmented RL methods, including SkillRL \citep{xia2026skillrl}, Skill0 \citep{lu2026skill0}, and D2Skill \citep{tu2026dynamic}. For search-augmented QA, we additionally compare with search-oriented reasoning baselines, including Search-o1 \citep{li2025search}, Search-R1 \citep{jin2025search}, ZeroSearch \citep{sun2025zerosearch}, and StepSearch \citep{zheng2025stepsearch}.

\noindent\textbf{Implementation Details.}
We use \texttt{Qwen2.5-7B-Instruct} and \texttt{Qwen3-4B-Instruct}~\citep{yang2025qwen3} as backbones.
For ALFWorld, we follow GiGPO~\citep{fenggroup}; for Search-QA, we follow
Search-R1~\citep{jin2025search} with E5 retrieval~\citep{wang2022text}, training
on NQ and HotpotQA and evaluating on the remaining QA datasets. The skill bank is
initialized from SkillRL~\citep{xia2026skillrl}. Unless otherwise specified, we set the induction-unit size $K=4$, promotion
ratio $\rho=20\%$, and novelty threshold $\gamma=0.8$. Full implementation details are in
Appendix~\ref{app:details}. A hyperparameter analysis of $K$ is provided in Appendix~\ref{app:hyperparameter}.
\begin{table*}[h]
\centering
\small
\setlength{\tabcolsep}{5pt}
\caption{Main results on ALFWorld and WebShop. \bestcap{} and \secondcap{} results are highlighted.}
\begin{tabular}{lccccccc|cc}
\toprule
\multirow{2}{*}{Method} & \multicolumn{7}{c|}{ALFWorld} & \multicolumn{2}{c}{WebShop} \\
\cmidrule(lr){2-8} \cmidrule(lr){9-10}
& Pick & Look & Clean & Heat & Cool & Pick2 & All & Score & Succ. \\
\midrule
\multicolumn{10}{l}{\textit{Closed-source LLMs}} \\
GPT-4o           & 75.3 & 60.8 & 31.2 & 56.7 & 21.6 & 49.8 & 48.0 & 31.8 & 23.7 \\
Gemini-2.5-Pro   & 92.8 & 63.3 & 62.1 & 69.0 & 26.6 & 58.7 & 60.3 & 42.5 & 35.9 \\
\midrule
\multicolumn{10}{l}{\textit{Qwen2.5-7B-Instruct}} \\
Qwen2.5          & 33.4 & 21.6 & 19.3 & 6.90 & 2.80 & 3.20 & 14.8 & 26.4 & 7.80 \\
\midrule
\multicolumn{10}{l}{\textit{Prompt-based Agentic or Memory-based Methods}} \\
ReAct$^{*}$      & 48.5 & 35.4 & 34.3 & 13.2 & 18.2 & 17.6 & 31.2 & 46.2 & 19.5 \\
Reflexion$^{*}$  & 62.0 & 41.6 & 44.9 & 30.9 & 36.3 & 23.8 & 42.7 & 58.1 & 28.8 \\
Mem0             & 54.0 & 55.0 & 26.9 & 36.4 & 20.8 & 7.69 & 33.6 & 23.9 & 2.00 \\
ExpeL            & 21.0 & 67.0 & 55.0 & 52.0 & 71.0 & 6.00 & 46.3 & 30.9 & 11.2 \\
MemP             & 54.3 & 38.5 & 48.1 & 56.2 & 32.0 & 16.7 & 41.4 & 25.3 & 6.40 \\
SimpleMem        & 64.5 & 33.3 & 20.0 & 12.5 & 33.3 & 3.84 & 29.7 & 33.2 & 8.59 \\
\midrule
\multicolumn{10}{l}{\textit{RL-based Methods}} \\
RLOO$^{*}$       & 87.6 & 78.2 & 87.3 & 81.3 & 71.9 & 48.9 & 75.5 & 80.3 & 65.7 \\
GRPO$^{*}$       & 90.8 & 66.1 & 89.3 & 74.7 & 72.5 & 64.7 & 77.6 & 79.3 & 66.1 \\
\midrule
\multicolumn{10}{l}{\textit{Memory-Augmented RL-based Methods}} \\
MemRL            & 62.8 & 38.5 & 22.2 & 12.5 & 8.00 & 0.00 & 21.4 & 29.5 & 9.20 \\
EvolveR          & 64.9 & 33.3 & 46.4 & 13.3 & 33.3 & 33.3 & 43.8 & 42.5 & 17.6 \\
Mem0+GRPO        & 78.1 & 54.8 & 56.1 & 31.0 & 65.0 & 26.9 & 54.7 & 58.1 & 37.5 \\
SimpleMem+GRPO   & 89.5 & 36.3 & 60.0 & 50.0 & 64.9 & 26.3 & 62.5 & 67.8 & 46.9 \\
SKILLRL          & \second{97.9} & 71.4 & 90.0 & \second{90.0} & \best{95.5} & \second{87.5} & \second{89.9} & \second{85.2} & 72.7 \\
Skill0           & 95.6 & \second{80.4} & \best{100} & 86.7 & 78.7 & 75.2 & 87.9 & 83.2 & 71.9 \\
D2Skill          & 93.8 & \best{94.7} & 95.5 & 77.8 & \second{95.0} & 72.0 & 87.8 & 83.4 & \second{73.4} \\
SAPO             & \best{98.7} & 73.9 & \second{98.1} & \best{92.6} & 85.0 & \best{89.2} & \best{92.2} & \best{90.5} & \best{78.1} \\
\bottomrule
\end{tabular}
\label{tab:alfworld_webshop}
\end{table*}
\subsection{Main Results}
\paragraph{Results on ALFWorld and WebShop.}
As shown in Table~\ref{tab:alfworld_webshop}, SAPO achieves the best overall performance among the compared methods. The consistent improvement over SkillRL, Skill0, and D2Skill indicates that performance in skill-augmented RL depends not only on expanding the skill bank, but also on controlling the quality of the skills that enter it. Appendix~\ref{app:qwen3_4b_results} shows similar gains when using
\texttt{Qwen3-4B-Instruct} as the base model.

\paragraph{Results on Search-Augmented QA.}
As shown in Table~\ref{tab:single_multi_hop_qa}, SAPO achieves the best average
performance among the compared methods. Its gains on both training-domain and
out-of-domain QA benchmarks suggest that marginal-utility feedback helps the
skill generator learn transferable search and reasoning strategies rather than
dataset-specific shortcuts.

\begin{table*}[h]
\centering
\small
\setlength{\tabcolsep}{6pt}
\caption{Results on single-hop and multi-hop QA benchmarks. $\dagger$ and $\star$ indicate in-domain and out-of-domain datasets, respectively.}
\begin{tabular}{lcccccccc}
\toprule
\multirow{2}{*}{Method} & \multicolumn{3}{c}{Single-Hop QA} & \multicolumn{4}{c}{Multi-Hop QA} & \multirow{2}{*}{Avg.} \\
\cmidrule(lr){2-4} \cmidrule(lr){5-8}
& NQ$^{\dagger}$ & TriviaQA$^{*}$ & PopQA$^{*}$ & HotpotQA$^{\dagger}$ & 2Wiki$^{*}$ & MuSiQue$^{*}$ & Bamboogle$^{*}$ & \\
\midrule
\multicolumn{9}{l}{\textit{Qwen2.5-7B-Instruct}} \\
Qwen2.5   & 11.6 & 35.6 & 1.20 & 16.4 & 22.2 & 4.80 & 14.4 & 15.2 \\
CoT       & 12.8 & 35.6 & 3.80 & 16.2 & 22.6 & 6.60 & 24.0 & 17.4 \\
RAG       & 27.4 & 58.2 & 17.8 & 25.8 & 23.2 & 9.40 & 16.8 & 25.5 \\
Search-o1 & 19.4 & 40.6 & 11.4 & 17.0 & 27.0 & 8.60 & 30.4 & 22.1 \\
R1-Instruct     & 21.0 & 44.9 & 17.1 & 20.8 & 27.5 & 6.00 & 19.2 & 22.4 \\
Search-R1       & 39.3 & 61.0 & 39.7 & 37.0 & 40.1 & 14.6 & 36.8 & 38.5 \\
ZeroSearch      & 43.6 & 61.8 & \best{51.5} & 34.6 & 35.2 & \second{18.4} & 27.8 & 39.1 \\
StepSearch      & --   & --   & --   & 38.6 & 36.6 & \best{22.6} & 40.0 & --   \\
EvolveR         & 43.5 & \second{63.4} & 44.6 & 38.2 & \second{42.0} & 15.6 & \second{54.4} & 43.1 \\
\midrule
SKILLRL & \second{45.9} & 61.2 & 45.9 & \second{43.9} & 39.9 & 17.2 & 45.6 & \second{45.5} \\
Skill0    & 42.7 & \best{63.6} & 45.3 & 40.0 & 38.3 & 16.4 & \best{66.9} & 44.4 \\
SAPO             & \best{48.4} & 62.9 & \second{46.7} & \best{45.0} & \best{45.2} & 18.3 & 46.4 & \best{47.8} \\
\bottomrule
\end{tabular}
\label{tab:single_multi_hop_qa}
\end{table*}

\subsection{Training Curves}
\begin{figure}[h]
    \centering
    \includegraphics[width=0.75\linewidth]{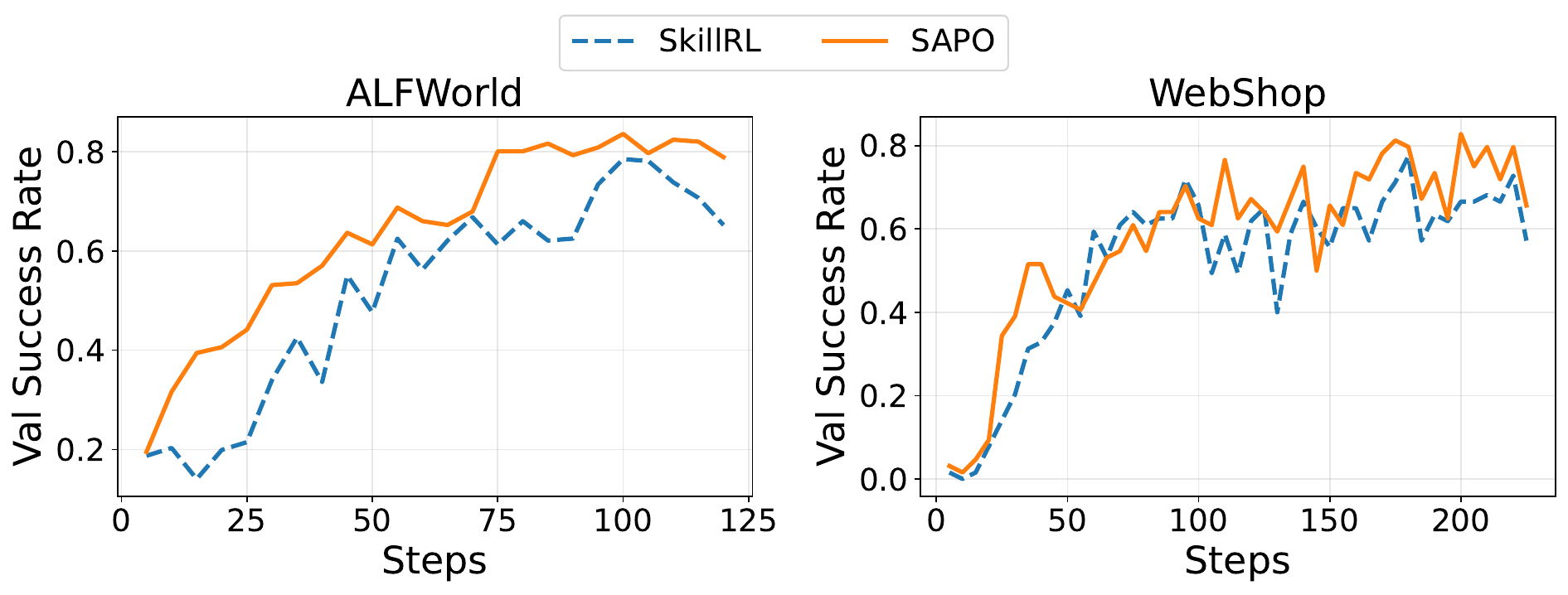}
    \caption{Validation success rate of SAPO and SkillRL on ALFWorld and WebShop.}
    \label{fig:curves}
\end{figure}

We compare the training dynamics of SAPO and SkillRL~\citep{xia2026skillrl}
using validation success rate, with \texttt{Qwen3-4B-Instruct} as the base model. As shown in Figure~\ref{fig:curves}, we make two observations.
\textbf{(1)} SAPO improves faster than SkillRL on both benchmarks, suggesting
that utility-based skill filtering provides more reliable guidance when
the base policy is still weak.
\textbf{(2)} SAPO achieves higher best validation performance, consistent with
the main results and indicating that its gains generalize across backbones.
Appendix~\ref{app:alfworld_subtask_dynamics} provides training curves on individual ALFWorld subtasks.

\subsection{Skill Utility}
\begin{figure}
    \centering
    \includegraphics[width=0.75\linewidth]{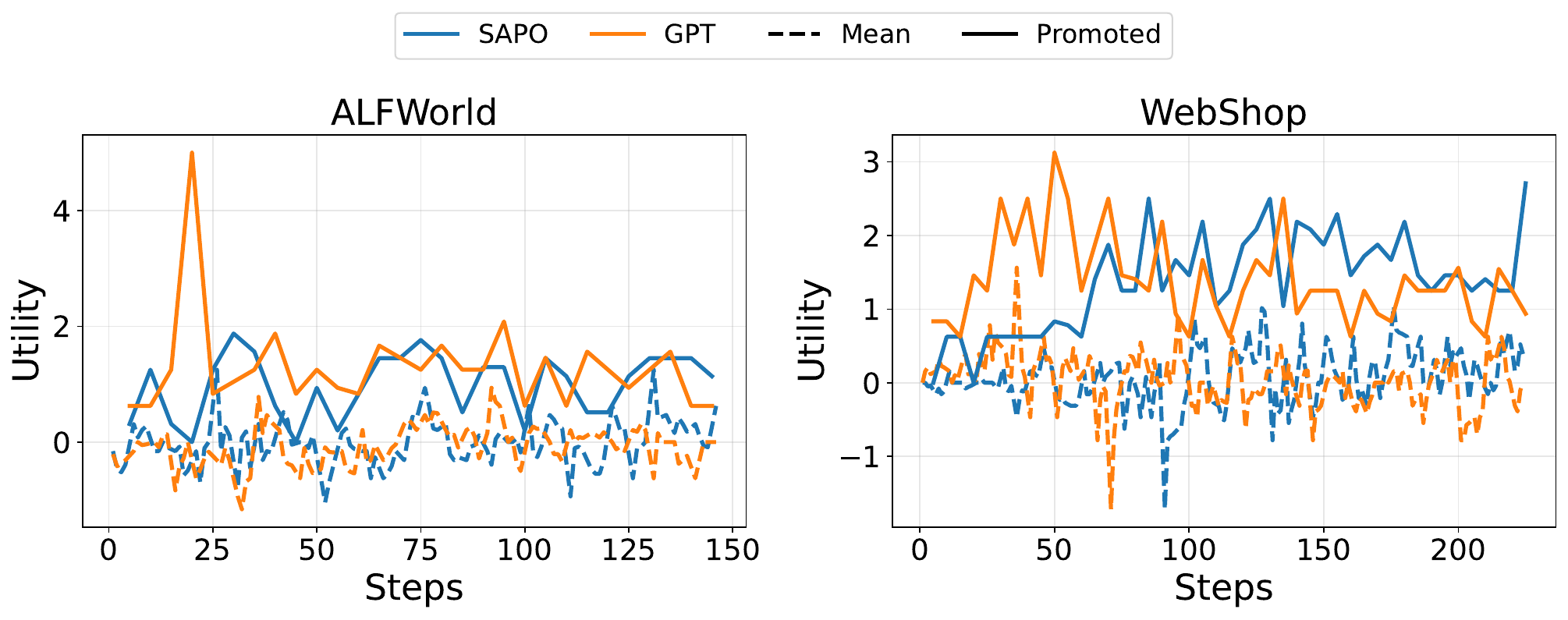}
    \caption{Skill utility of SAPO and SkillRL on ALFWorld and WebShop during training.}
    \label{fig:utility}
\end{figure}

In this section, we analyze the training dynamics of generated skill utility. We compare SAPO with SkillRL, where SAPO uses \texttt{Qwen3-4B-Instruct} as the skill generator, while SkillRL uses GPT-5.4. The results are shown in Figure~\ref{fig:utility}.
From the figure, we make two observations. 
\textbf{(1)} In the early stage of training, SAPO generates skills with lower utility than those generated by GPT. This is expected because SAPO relies on a smaller open-weight model as its skill generator. However, as training progresses, the SAPO skill generator learns to produce more beneficial skills. Its skill utility becomes comparable to, and in some cases slightly better than, that of the GPT-based generator used in SkillRL. This demonstrates the benefit of training the skill generator with utility feedback.
\textbf{(2)} Even when the policy becomes stronger in the later stage of training, the SAPO skill generator can still consistently produce skills with positive utility. This suggests that useful skills can continue to provide additional gains beyond an improved base policy.
\label{sec:skill_utility}

\subsection{Ablation Study}
\begin{table}[h]
\centering
\small
\setlength{\tabcolsep}{6pt}
\caption{Ablation study on ALFWorld and WebShop.}
\begin{tabular}{lccc}
\toprule
\multirow{2}{*}{Method} & ALFWorld & \multicolumn{2}{c}{WebShop} \\
\cmidrule(lr){2-2} \cmidrule(lr){3-4}
& All. & Score & Succ. \\
\midrule
SAPO w/o Validation & 90.6 & 83.0 & 75.0 \\
SAPO w/o Generator  & 90.2 & 83.2 & 73.4 \\
SAPO w/o Scoring    & 91.4 & 86.5 & 76.6  \\
\midrule
SAPO                & 92.2 & 90.5 & 78.1 \\
\bottomrule
\end{tabular}
\label{tab:ablation}
\end{table}
We conduct ablation studies to evaluate the contribution of each major component
of SAPO. Specifically, we compare SAPO with three variants: \textbf{SAPO w/o
Validation}, which removes utility-based skill promotion and directly stores
newly generated skills; \textbf{SAPO w/o Generator}, which removes
utility-weighted skill generator training; and \textbf{SAPO w/o Scoring}, which
removes likelihood-based skill pruning and retrieval-time reranking. Table~\ref{tab:ablation}
reports the results on ALFWorld and WebShop.
From the table, we make the following observations.
\textbf{(1)} Utility-based validation improves skill-bank quality. Removing it
degrades performance on both benchmarks, showing that directly storing generated
skills can introduce low-quality skills.
\textbf{(2)} Utility-weighted generator training improves skill generation.
Without it, performance drops, especially on WebShop success rate, indicating
that utility feedback helps the policy generate more useful skills.
\textbf{(3)} Likelihood-based scoring improves skill reuse. Removing pruning and
retrieval-time reranking lowers performance, suggesting that the policy's
skill-generation likelihood helps select useful skills and remove outdated ones.
Overall, SAPO achieves the best results across all metrics.

\section{Conclusion}
We presented SAPO, an online RL framework that validates generated skills before
storage and uses marginal utility to improve skill generation, maintenance, and
retrieval. SAPO derives skill-utility signals from the same rollouts used for
agent learning, enabling skill generator training without extra
rollout cost or repeated proprietary LLM calls. Experiments show consistent gains
over prior skill-augmented RL methods across interactive decision-making and
search-augmented QA tasks.

{
\newpage
\bibliographystyle{unsrt}
\bibliography{ref}

@article{wang2022text,
  title={Text embeddings by weakly-supervised contrastive pre-training},
  author={Wang, Liang and Yang, Nan and Huang, Xiaolong and Jiao, Binxing and Yang, Linjun and Jiang, Daxin and Majumder, Rangan and Wei, Furu},
  journal={arXiv preprint arXiv:2212.03533},
  year={2022}
}

@article{fenggroup,
  title={Group-in-group policy optimization for llm agent training},
  author={Feng, Lang and Xue, Zhenghai and Liu, Tingcong and An, Bo},
  journal={arXiv preprint arXiv:2505.10978},
  year={2025}
}

@article{jin2025search,
  title={Search-r1: Training llms to reason and leverage search engines with reinforcement learning},
  author={Jin, Bowen and Zeng, Hansi and Yue, Zhenrui and Yoon, Jinsung and Arik, Sercan and Wang, Dong and Zamani, Hamed and Han, Jiawei},
  journal={arXiv preprint arXiv:2503.09516},
  year={2025}
}

@article{zheng2025stepsearch,
  title={Stepsearch: Igniting llms search ability via step-wise proximal policy optimization},
  author={Wang, Ziliang and Zheng, Xuhui and An, Kang and Ouyang, Cijun and Cai, Jialu and Wang, Yuhang and Wu, Yichao},
  journal={arXiv preprint arXiv:2505.15107},
  year={2025}
}

@article{sun2025zerosearch,
  title={Zerosearch: Incentivize the search capability of llms without searching},
  author={Sun, Hao and Qiao, Zile and Guo, Jiayan and Fan, Xuanbo and Hou, Yingyan and Jiang, Yong and Xie, Pengjun and Zhang, Yan and Huang, Fei and Zhou, Jingren},
  journal={arXiv preprint arXiv:2505.04588},
  year={2025}
}

@article{yang2025qwen3,
  title={Qwen3 technical report},
  author={Yang, An and Li, Anfeng and Yang, Baosong and Zhang, Beichen and Hui, Binyuan and Zheng, Bo and Yu, Bowen and Gao, Chang and Huang, Chengen and Lv, Chenxu and others},
  journal={arXiv preprint arXiv:2505.09388},
  year={2025}
}

@article{zhou2026memento,
  title={Memento-skills: Let agents design agents},
  author={Zhou, Huichi and Guo, Siyuan and Liu, Anjie and Yu, Zhongwei and Gong, Ziqin and Zhao, Bowen and Chen, Zhixun and Zhang, Menglong and Chen, Yihang and Li, Jinsong and others},
  journal={arXiv preprint arXiv:2603.18743},
  year={2026}
}

@inproceedings{li2025search,
  title={Search-o1: Agentic search-enhanced large reasoning models},
  author={Li, Xiaoxi and Dong, Guanting and Jin, Jiajie and Zhang, Yuyao and Zhou, Yujia and Zhu, Yutao and Zhang, Peitian and Dou, Zhicheng},
  booktitle={Proceedings of the 2025 Conference on Empirical Methods in Natural Language Processing},
  pages={5420--5438},
  year={2025}
}

@article{liu2026simplemem,
  title={SimpleMem: Efficient Lifelong Memory for LLM Agents},
  author={Liu, Jiaqi and Su, Yaofeng and Xia, Peng and Han, Siwei and Zheng, Zeyu and Xie, Cihang and Ding, Mingyu and Yao, Huaxiu},
  journal={arXiv preprint arXiv:2601.02553},
  year={2026}
}

@inproceedings{zhao2024expel,
  title={Expel: Llm agents are experiential learners},
  author={Zhao, Andrew and Huang, Daniel and Xu, Quentin and Lin, Matthieu and Liu, Yong-Jin and Huang, Gao},
  booktitle={Proceedings of the AAAI Conference on Artificial Intelligence},
  volume={38},
  pages={19632--19642},
  year={2024}
}

@article{zhang2026memrl,
  title={Memrl: Self-evolving agents via runtime reinforcement learning on episodic memory},
  author={Zhang, Shengtao and Wang, Jiaqian and Zhou, Ruiwen and Liao, Junwei and Feng, Yuchen and Li, Zhuo and Zheng, Yujie and Zhang, Weinan and Wen, Ying and Li, Zhiyu and others},
  journal={arXiv preprint arXiv:2601.03192},
  year={2026}
}

@inproceedings{ahmadian2024back,
  title={Back to basics: Revisiting REINFORCE-style optimization for learning from human feedback in LLMs},
  author={Ahmadian, Arash and Cremer, Chris and Gall{\'e}, Matthias and Fadaee, Marzieh and Kreutzer, Julia and Pietquin, Olivier and {\"U}st{\"u}n, Ahmet and Hooker, Sara},
  booktitle={Proceedings of the 62nd Annual Meeting of the Association for Computational Linguistics (Volume 1: Long Papers)},
  pages={12248--12267},
  year={2024}
}

@article{fang2025memp,
  title={Memp: Exploring agent procedural memory},
  author={Fang, Runnan and Liang, Yuan and Wang, Xiaobin and Wu, Jialong and Qiao, Shuofei and Xie, Pengjun and Huang, Fei and Chen, Huajun and Zhang, Ningyu},
  journal={arXiv preprint arXiv:2508.06433},
  year={2025}
}

@article{chhikara2025mem0,
  title={Mem0: Building production-ready ai agents with scalable long-term memory},
  author={Chhikara, Prateek and Khant, Dev and Aryan, Saket and Singh, Taranjeet and Yadav, Deshraj},
  journal={arXiv preprint arXiv:2504.19413},
  year={2025}
}

@article{yao2022react,
  title={React: Synergizing reasoning and acting in language models},
  author={Yao, Shunyu and Zhao, Jeffrey and Yu, Dian and Du, Nan and Shafran, Izhak and Narasimhan, Karthik and Cao, Yuan},
  journal={arXiv preprint arXiv:2210.03629},
  year={2022}
}

@article{comanici2025gemini,
  title={Gemini 2.5: Pushing the frontier with advanced reasoning, multimodality, long context, and next generation agentic capabilities},
  author={Comanici, Gheorghe and Bieber, Eric and Schaekermann, Mike and Pasupat, Ice and Sachdeva, Noveen and Dhillon, Inderjit and Blistein, Marcel and Ram, Ori and Zhang, Dan and Rosen, Evan and others},
  journal={arXiv preprint arXiv:2507.06261},
  year={2025}
}

@inproceedings{press2023measuring,
  title={Measuring and narrowing the compositionality gap in language models},
  author={Press, Ofir and Zhang, Muru and Min, Sewon and Schmidt, Ludwig and Smith, Noah A and Lewis, Mike},
  booktitle={Findings of the Association for Computational Linguistics: EMNLP 2023},
  pages={5687--5711},
  year={2023}
}

@inproceedings{ho2020constructing,
  title={Constructing a multi-hop qa dataset for comprehensive evaluation of reasoning steps},
  author={Ho, Xanh and Nguyen, Anh-Khoa Duong and Sugawara, Saku and Aizawa, Akiko},
  booktitle={Proceedings of the 28th International Conference on Computational Linguistics},
  pages={6609--6625},
  year={2020}
}

@article{trivedi2022musique,
  title={MuSiQue: Multi-hop Questions via Single-hop Question Composition},
  author={Trivedi, Harsh and Balasubramanian, Niranjan and Khot, Tushar and Sabharwal, Ashish},
  journal={Transactions of the Association for Computational Linguistics},
  volume={10},
  pages={539--554},
  year={2022}
}

@inproceedings{yang2018hotpotqa,
  title={HotpotQA: A dataset for diverse, explainable multi-hop question answering},
  author={Yang, Zhilin and Qi, Peng and Zhang, Saizheng and Bengio, Yoshua and Cohen, William and Salakhutdinov, Ruslan and Manning, Christopher D},
  booktitle={Proceedings of the 2018 conference on empirical methods in natural language processing},
  pages={2369--2380},
  year={2018}
}

@inproceedings{mallen2023not,
  title={When not to trust language models: Investigating effectiveness of parametric and non-parametric memories},
  author={Mallen, Alex and Asai, Akari and Zhong, Victor and Das, Rajarshi and Khashabi, Daniel and Hajishirzi, Hannaneh},
  booktitle={Proceedings of the 61st annual meeting of the association for computational linguistics (volume 1: Long papers)},
  pages={9802--9822},
  year={2023}
}

@inproceedings{joshi2017triviaqa,
  title={Triviaqa: A large scale distantly supervised challenge dataset for reading comprehension},
  author={Joshi, Mandar and Choi, Eunsol and Weld, Daniel S and Zettlemoyer, Luke},
  booktitle={Proceedings of the 55th Annual Meeting of the Association for Computational Linguistics (Volume 1: Long Papers)},
  pages={1601--1611},
  year={2017}
}

@article{kwiatkowski2019natural,
  title={Natural questions: a benchmark for question answering research},
  author={Kwiatkowski, Tom and Palomaki, Jennimaria and Redfield, Olivia and Collins, Michael and Parikh, Ankur and Alberti, Chris and Epstein, Danielle and Polosukhin, Illia and Devlin, Jacob and Lee, Kenton and others},
  journal={Transactions of the Association for Computational Linguistics},
  volume={7},
  pages={453--466},
  year={2019},
  publisher={MIT Press One Rogers Street, Cambridge, MA 02142-1209, USA journals-info~…}
}

@article{lu2026skill0,
  title={Skill0: In-context agentic reinforcement learning for skill internalization},
  author={Lu, Zhengxi and Yao, Zhiyuan and Wu, Jinyang and Han, Chengcheng and Gu, Qi and Cai, Xunliang and Lu, Weiming and Xiao, Jun and Zhuang, Yueting and Shen, Yongliang},
  journal={arXiv preprint arXiv:2604.02268},
  year={2026}
}

@article{zhou2508memento,
  title={Memento: Fine-tuning llm agents without fine-tuning llms},
  author={Zhou, Huichi and Chen, Yihang and Guo, Siyuan and Yan, Xue and Lee, Kin Hei and Wang, Zihan and Lee, Ka Yiu and Zhang, Guchun and Shao, Kun and Yang, Linyi and others},
  journal={arXiv preprint arXiv:2508.16153},
  year={2025}
}

@article{tu2026dynamic,
  title={Dynamic Dual-Granularity Skill Bank for Agentic RL},
  author={Tu, Songjun and Xu, Chengdong and Zhang, Qichao and Zhang, Yaocheng and Lan, Xiangyuan and Li, Linjing and Zhao, Dongbin},
  journal={arXiv preprint arXiv:2603.28716},
  year={2026}
}

@article{yao2022webshop,
  title={Webshop: Towards scalable real-world web interaction with grounded language agents},
  author={Yao, Shunyu and Chen, Howard and Yang, John and Narasimhan, Karthik},
  journal={Advances in Neural Information Processing Systems},
  volume={35},
  pages={20744--20757},
  year={2022}
}

@article{shridhar2020alfworld,
  title={Alfworld: Aligning text and embodied environments for interactive learning},
  author={Shridhar, Mohit and Yuan, Xingdi and C{\^o}t{\'e}, Marc-Alexandre and Bisk, Yonatan and Trischler, Adam and Hausknecht, Matthew},
  journal={arXiv preprint arXiv:2010.03768},
  year={2020}
}

@article{zhusurprising,
  title={The surprising effectiveness of negative reinforcement in llm reasoning},
  author={Zhu, Xinyu and Xia, Mengzhou and Wei, Zhepei and Chen, Wei-Lin and Chen, Danqi and Meng, Yu},
  journal={arXiv preprint arXiv:2506.01347},
  year={2025}
}

@article{shao2024deepseekmath,
  title={Deepseekmath: Pushing the limits of mathematical reasoning in open language models},
  author={Shao, Zhihong and Wang, Peiyi and Zhu, Qihao and Xu, Runxin and Song, Junxiao and Bi, Xiao and Zhang, Haowei and Zhang, Mingchuan and Li, YK and Wu, Yang and others},
  journal={arXiv preprint arXiv:2402.03300},
  year={2024}
}

@article{wang2025reinforcement,
  title={Reinforcement learning for self-improving agent with skill library},
  author={Wang, Jiongxiao and Yan, Qiaojing and Wang, Yawei and Tian, Yijun and Mishra, Soumya Smruti and Xu, Zhichao and Gandhi, Megha and Xu, Panpan and Cheong, Lin Lee},
  journal={arXiv preprint arXiv:2512.17102},
  year={2025}
}

@article{ye2026meta,
  title={Meta Context Engineering via Agentic Skill Evolution},
  author={Ye, Haoran and He, Xuning and Arak, Vincent and Dong, Haonan and Song, Guojie},
  journal={arXiv preprint arXiv:2601.21557},
  year={2026}
}

@article{zhang2026equipping,
  title={Equipping agents for the real world with Agent Skills, October 2025},
  author={Zhang, Barry and Lazuka, Keith and Murag, Mahesh},
  journal={URL https://www.anthropic.com/engineering/equipping-agents-for-the-real-world-with-agent-skills.Accessed},
  pages={01--28},
  year={2026}
}

@inproceedings{liu2024skillact,
  title={Skillact: Using skill abstractions improves llm agents},
  author={Liu, Anthony Zhe and Choi, Jongwook and Sohn, Sungryull and Fu, Yao and Kim, Jaekyeom and Kim, Dong-Ki and Wang, Xinhe and Yoo, Jaewon and Lee, Honglak},
  booktitle={ICML 2024 Workshop on LLMs and Cognition},
  year={2024}
}

@article{wang2025asi,
  title={Inducing programmatic skills for agentic tasks},
  author={Wang, Zora Zhiruo and Gandhi, Apurva and Neubig, Graham and Fried, Daniel},
  journal={arXiv preprint arXiv:2504.06821},
  year={2025}
}

@article{huang2025cascade,
  title={Cascade: Cumulative agentic skill creation through autonomous development and evolution},
  author={Huang, Xu and Chen, Junwu and Fei, Yuxing and Li, Zhuohan and Schwaller, Philippe and Ceder, Gerbrand},
  journal={arXiv preprint arXiv:2512.23880},
  year={2025}
}

@article{yang2025exif,
  title={Automated skill discovery for language agents through exploration and iterative feedback},
  author={Yang, Yongjin and Kang, Sinjae and Lee, Juyong and Lee, Dongjun and Yun, Se-Young and Lee, Kimin},
  journal={arXiv preprint arXiv:2506.04287},
  year={2025}
}

@inproceedings{zhou2025pae,
  title={Proposer-agent-evaluator (pae): Autonomous skill discovery for foundation model internet agents},
  author={Zhou, Yifei and Yang, Qianlan and Lin, Kaixiang and Bai, Min and Zhou, Xiong and Wang, Yu-Xiong and Levine, Sergey and Li, Li Erran},
  booktitle={Forty-second International Conference on Machine Learning},
  year={2025}
}

@article{wu2026survey,
  title={Agent Skills from the Perspective of Procedural Memory: A Survey},
  author={Wu, Yaxiong and Zhang, Yongyue},
  journal={Authorea Preprints},
  year={2026},
  publisher={Authorea}
}

@article{shinn2023reflexion,
  title={Reflexion: Language agents with verbal reinforcement learning},
  author={Shinn, Noah and Cassano, Federico and Gopinath, Ashwin and Narasimhan, Karthik and Yao, Shunyu},
  journal={Advances in neural information processing systems},
  volume={36},
  pages={8634--8652},
  year={2023}
}

@article{suttonWelcomeEraExperience2025,
  title={Welcome to the era of experience},
  author={Silver, David and Sutton, Richard S},
  journal={Google AI},
  volume={1},
  pages={11},
  year={2025}
}

@article{ouyang2025reasoningbankscalingagentselfevolving,
  title={Reasoningbank: Scaling agent self-evolving with reasoning memory},
  author={Ouyang, Siru and Yan, Jun and Hsu, I and Chen, Yanfei and Jiang, Ke and Wang, Zifeng and Han, Rujun and Le, Long T and Daruki, Samira and Tang, Xiangru and others},
  journal={arXiv preprint arXiv:2509.25140},
  year={2025}
}

@inproceedings{huang_r2d2_2025,
  title={R2d2: Remembering, replaying and dynamic decision making with a reflective agentic memory},
  author={Huang, Tenghao and Basu, Kinjal and Abdelaziz, Ibrahim and Kapanipathi, Pavan and May, Jonathan and Chen, Muhao},
  booktitle={Proceedings of the 63rd Annual Meeting of the Association for Computational Linguistics (Volume 1: Long Papers)},
  pages={30318--30330},
  year={2025}
}

@article{wang_voyager_2024,
  title={Voyager: An open-ended embodied agent with large language models},
  author={Wang, Guanzhi and Xie, Yuqi and Jiang, Yunfan and Mandlekar, Ajay and Xiao, Chaowei and Zhu, Yuke and Fan, Linxi and Anandkumar, Anima},
  journal={arXiv preprint arXiv:2305.16291},
  year={2023}
}

@article{skillweaver,
  title={Skillweaver: Web agents can self-improve by discovering and honing skills},
  author={Zheng, Boyuan and Fatemi, Michael Y and Jin, Xiaolong and Wang, Zora Zhiruo and Gandhi, Apurva and Song, Yueqi and Gu, Yu and Srinivasa, Jayanth and Liu, Gaowen and Neubig, Graham and others},
  journal={arXiv preprint arXiv:2504.07079},
  year={2025}
}

@article{tang2025agentkbleveragingcrossdomain,
  title={Agent kb: Leveraging cross-domain experience for agentic problem solving},
  author={Tang, Xiangru and Qin, Tianrui and Peng, Tianhao and Zhou, Ziyang and Shao, Daniel and Du, Tingting and Wei, Xinming and Xia, Peng and Wu, Fang and Zhu, He and others},
  journal={arXiv preprint arXiv:2507.06229},
  year={2025}
}

@article{chen2025scalingagentlearningexperience,
  title={Scaling agent learning via experience synthesis},
  author={Chen, Zhaorun and Zhao, Zhuokai and Zhang, Kai and Liu, Bo and Qi, Qi and Wu, Yifan and Kalluri, Tarun and Cao, Sara and Xiong, Yuanhao and Tong, Haibo and others},
  journal={arXiv preprint arXiv:2511.03773},
  year={2025}
}

@article{agent-early-experience,
  title={Agent learning via early experience},
  author={Zhang, Kai and Chen, Xiangchao and Liu, Bo and Xue, Tianci and Liao, Zeyi and Liu, Zhihan and Wang, Xiyao and Ning, Yuting and Chen, Zhaorun and Fu, Xiaohan and others},
  journal={arXiv preprint arXiv:2510.08558},
  year={2025}
}

@inproceedings{suzgun2025dynamiccheatsheettesttimelearning,
  title={Dynamic cheatsheet: Test-time learning with adaptive memory},
  author={Suzgun, Mirac and Yuksekgonul, Mert and Bianchi, Federico and Jurafsky, Dan and Zou, James},
  booktitle={Proceedings of the 19th Conference of the European Chapter of the Association for Computational Linguistics (Volume 1: Long Papers)},
  pages={7080--7106},
  year={2026}
}

@article{cai2025flexcontinuousagentevolution,
  title={Flex: Continuous agent evolution via forward learning from experience},
  author={Cai, Zhicheng and Guo, Xinyuan and Pei, Yu and Feng, Jiangtao and Su, Jinsong and Chen, Jiangjie and Zhang, Ya-Qin and Ma, Wei-Ying and Wang, Mingxuan and Zhou, Hao},
  journal={arXiv preprint arXiv:2511.06449},
  year={2025}
}

@article{zhang_darwin_2025,
  title={Darwin godel machine: Open-ended evolution of self-improving agents},
  author={Zhang, Jenny and Hu, Shengran and Lu, Cong and Lange, Robert and Clune, Jeff},
  journal={arXiv preprint arXiv:2505.22954},
  year={2025}
}

@article{han_legomem_2025,
  title={Legomem: Modular procedural memory for multi-agent llm systems for workflow automation},
  author={Han, Dongge and Couturier, Camille and Diaz, Daniel Madrigal and Zhang, Xuchao and R{\"u}hle, Victor and Rajmohan, Saravan},
  journal={arXiv preprint arXiv:2510.04851},
  year={2025}
}

@article{Zhang2025GMemory,
  title={G-memory: Tracing hierarchical memory for multi-agent systems},
  author={Zhang, Guibin and Fu, Muxin and Wan, Guancheng and Yu, Miao and Wang, Kun and Yan, Shuicheng},
  journal={arXiv preprint arXiv:2506.07398},
  year={2025}
}

@article{wu2025evolverselfevolvingllmagents,
  title={Evolver: Self-evolving llm agents through an experience-driven lifecycle},
  author={Wu, Rong and Wang, Xiaoman and Mei, Jianbiao and Cai, Pinlong and Fu, Daocheng and Yang, Cheng and Wen, Licheng and Yang, Xuemeng and Shen, Yufan and Wang, Yuxin and others},
  journal={arXiv preprint arXiv:2510.16079},
  year={2025}
}

@article{zhang2025memevolvemetaevolutionagentmemory,
  title={Memevolve: Meta-evolution of agent memory systems},
  author={Zhang, Guibin and Ren, Haotian and Zhan, Chong and Zhou, Zhenhong and Wang, Junhao and Zhu, He and Zhou, Wangchunshu and Yan, Shuicheng},
  journal={arXiv preprint arXiv:2512.18746},
  year={2025}
}

@article{zhai2025agentevolverefficientselfevolvingagent,
  title={Agentevolver: Towards efficient self-evolving agent system},
  author={Zhai, Yunpeng and Tao, Shuchang and Chen, Cheng and Zou, Anni and Chen, Ziqian and Fu, Qingxu and Mai, Shinji and Yu, Li and Deng, Jiaji and Cao, Zouying and others},
  journal={arXiv preprint arXiv:2511.10395},
  year={2025}
}

@article{cai2025experiencedriven,
  title={Building self-evolving agents via experience-driven lifelong learning: A framework and benchmark},
  author={Cai, Yuxuan and Hao, Yipeng and Zhou, Jie and Yan, Hang and Lei, Zhikai and Zhen, Rui and Han, Zhenhua and Yang, Yutao and Li, Junsong and Pan, Qianjun and others},
  journal={arXiv preprint arXiv:2508.19005},
  year={2025}
}

@article{yan2025memory,
  title={Memory-r1: Enhancing large language model agents to manage and utilize memories via reinforcement learning},
  author={Yan, Sikuan and Yang, Xiufeng and Huang, Zuchao and Nie, Ercong and Ding, Zifeng and Li, Zonggen and Ma, Xiaowen and Bi, Jinhe and Kersting, Kristian and Pan, Jeff Z and others},
  journal={arXiv preprint arXiv:2508.19828},
  year={2025}
}

@article{xia2026skillrl,
  title={Skillrl: Evolving agents via recursive skill-augmented reinforcement learning},
  author={Xia, Peng and Chen, Jianwen and Wang, Hanyang and Liu, Jiaqi and Zeng, Kaide and Wang, Yu and Han, Siwei and Zhou, Yiyang and Zhao, Xujiang and Chen, Haifeng and others},
  journal={arXiv preprint arXiv:2602.08234},
  year={2026}
}

@article{cao2025reme,
  title={Remember me, refine me: A dynamic procedural memory framework for experience-driven agent evolution},
  author={Cao, Zouying and Deng, Jiaji and Yu, Li and Zhou, Weikang and Liu, Zhaoyang and Ding, Bolin and Zhao, Hai},
  journal={arXiv preprint arXiv:2512.10696},
  year={2025}
}

@article{jiang2026sok,
  title={SoK: Agentic Skills--Beyond Tool Use in LLM Agents},
  author={Jiang, Yanna and Li, Delong and Deng, Haiyu and Ma, Baihe and Wang, Xu and Wang, Qin and Yu, Guangsheng},
  journal={arXiv preprint arXiv:2602.20867},
  year={2026}
}

@article{xu2026agentskills,
  title={Agent skills for large language models: Architecture, acquisition, security, and the path forward},
  author={Xu, Renjun and Yan, Yang},
  journal={arXiv preprint arXiv:2602.12430},
  year={2026}
}

@article{zhou2026externalization,
  title={Externalization in LLM Agents: A Unified Review of Memory, Skills, Protocols and Harness Engineering},
  author={Zhou, Chenyu and Chai, Huacan and Chen, Wenteng and Guo, Zihan and Shan, Rong and Song, Yuanyi and Xu, Tianyi and Yang, Yingxuan and Yu, Aofan and Zhang, Weiming and others},
  journal={arXiv preprint arXiv:2604.08224},
  year={2026}
}

@article{zhang2026experiencecompression,
  title={Experience Compression Spectrum: Unifying Memory, Skills, and Rules in LLM Agents},
  author={Zhang, Xing and Wang, Guanghui and Cui, Yanwei and Qiu, Wei and Li, Ziyuan and Zhu, Bing and He, Peiyang},
  journal={arXiv preprint arXiv:2604.15877},
  year={2026}
}

@article{li2026skillsbench,
  title={SkillsBench: Benchmarking how well agent skills work across diverse tasks},
  author={Li, Xiangyi and Chen, Wenbo and Liu, Yimin and Zheng, Shenghan and Chen, Xiaokun and He, Yifeng and Li, Yubo and You, Bingran and Shen, Haotian and Sun, Jiankai and others},
  journal={arXiv preprint arXiv:2602.12670},
  year={2026}
}

@article{zhong2026skilllearnbench,
  title={SkillLearnBench: Benchmarking Continual Learning Methods for Agent Skill Generation on Real-World Tasks},
  author={Zhong, Shanshan and Lu, Yi and Ning, Jingjie and Wan, Yibing and Feng, Lihan and Ao, Yuyi and Ribeiro, Leonardo FR and Dreyer, Markus and Ammirati, Sean and Xiong, Chenyan},
  journal={arXiv preprint arXiv:2604.20087},
  year={2026}
}

@article{liu2026skillswild,
  title={How Well Do Agentic Skills Work in the Wild: Benchmarking LLM Skill Usage in Realistic Settings},
  author={Liu, Yujian and Ji, Jiabao and An, Li and Jaakkola, Tommi and Zhang, Yang and Chang, Shiyu},
  journal={arXiv preprint arXiv:2604.04323},
  year={2026}
}

@article{wang2026skilltester,
  title={SkillTester: Benchmarking Utility and Security of Agent Skills},
  author={Wang, Leye and Wang, Zixing and Xu, Anjie},
  journal={arXiv preprint arXiv:2603.28815},
  year={2026}
}

@article{ni2026trace2skill,
  title={Trace2skill: Distill trajectory-local lessons into transferable agent skills},
  author={Ni, Jingwei and Liu, Yihao and Liu, Xinpeng and Sun, Yutao and Zhou, Mengyu and Cheng, Pengyu and Wang, Dexin and Jiang, Xiaoxi and Jiang, Guanjun},
  journal={arXiv preprint arXiv:2603.25158},
  year={2026}
}

@article{zhang2026coevoskills,
  title={EvoSkills: Self-Evolving Agent Skills via Co-Evolutionary Verification},
  author={Zhang, Hanrong and Fan, Shicheng and Zou, Henry Peng and Chen, Yankai and Wang, Zhenting and Zhou, Jiayu and Li, Chengze and Huang, Wei-Chieh and Yao, Yifei and Zheng, Kening and others},
  journal={arXiv preprint arXiv:2604.01687},
  year={2026}
}

@article{ma2026skillclaw,
  title={SkillClaw: Let Skills Evolve Collectively with Agentic Evolver},
  author={Ma, Ziyu and Yang, Shidong and Ji, Yuxiang and Wang, Xucong and Wang, Yong and Hu, Yiming and Huang, Tongwen and Chu, Xiangxiang},
  journal={arXiv preprint arXiv:2604.08377},
  year={2026}
}

@article{li2026arise,
  title={Arise: Agent reasoning with intrinsic skill evolution in hierarchical reinforcement learning},
  author={Li, Yu and Miao, Rui and Qi, Zhengling and Lan, Tian},
  journal={arXiv preprint arXiv:2603.16060},
  year={2026}
}

@article{zhang2026skillflow,
  title={SkillFlow: Benchmarking Lifelong Skill Discovery and Evolution for Autonomous Agents},
  author={Zhang, Ziao and Shi, Kou and Huang, Shiting and Nie, Avery and Zeng, Yu and Zhao, Yiming and Fang, Zhen and Su, Qishen and Qiu, Haibo and Yang, Wei and others},
  journal={arXiv preprint arXiv:2604.17308},
  year={2026}
}

@article{yang2026autoskill,
  title={Autoskill: Experience-driven lifelong learning via skill self-evolution},
  author={Yang, Yutao and Li, Junsong and Pan, Qianjun and Zhan, Bihao and Cai, Yuxuan and Du, Lin and Zhou, Jie and Chen, Kai and Chen, Qin and Li, Xin and others},
  journal={arXiv preprint arXiv:2603.01145},
  year={2026}
}

@article{xu2026ael,
  title={AEL: Agent Evolving Learning for Open-Ended Environments},
  author={Xu, Wujiang and Han, Jiaojiao and Guo, Minghao and Mei, Kai and Zhu, Xi and Zhang, Han and Metaxas, Dimitris N},
  journal={arXiv preprint arXiv:2604.21725},
  year={2026}
}

@article{du2026memorysurvey,
  title={Memory for autonomous LLM agents: Mechanisms, evaluation, and emerging frontiers},
  author={Du, Pengfei},
  journal={arXiv preprint arXiv:2603.07670},
  year={2026}
}

@article{yang2026graphmemory,
  title={Graph-based Agent Memory: Taxonomy, Techniques, and Applications},
  author={Yang, Chang and Zhou, Chuang and Xiao, Yilin and Dong, Su and Zhuang, Luyao and Zhang, Yujing and Wang, Zhu and Hong, Zijin and Yuan, Zheng and Xiang, Zhishang and others},
  journal={arXiv preprint arXiv:2602.05665},
  year={2026}
}

@article{fang2026trajectorymemory,
  title={Trajectory-Informed Memory Generation for Self-Improving Agent Systems},
  author={Fang, Gaodan and Isahagian, Vatche and Jayaram, KR and Kumar, Ritesh and Muthusamy, Vinod and Oum, Punleuk and Thomas, Gegi},
  journal={arXiv preprint arXiv:2603.10600},
  year={2026}
}

@article{yu2026agemem,
  title={Agentic memory: Learning unified long-term and short-term memory management for large language model agents},
  author={Yu, Yi and Yao, Liuyi and Xie, Yuexiang and Tan, Qingquan and Feng, Jiaqi and Li, Yaliang and Wu, Libing},
  journal={arXiv preprint arXiv:2601.01885},
  year={2026}
}

@article{shen2026skillfoundry,
  title={SKILLFOUNDRY: Building Self-Evolving Agent Skill Libraries from Heterogeneous Scientific Resources},
  author={Shen, Shuaike and Cheng, Wenduo and Ma, Mingqian and Turcan, Alistair and Zhang, Martin Jinye and Ma, Jian},
  journal={arXiv preprint arXiv:2604.03964},
  year={2026}
}

@article{wang2026skillx,
  title={SkillX: Automatically constructing skill knowledge bases for agents},
  author={Wang, Chenxi and Yu, Zhuoyun and Xie, Xin and Yao, Wuguannan and Fang, Runnan and Qiao, Shuofei and Cao, Kexin and Zheng, Guozhou and Qi, Xiang and Zhang, Peng and others},
  journal={arXiv preprint arXiv:2604.04804},
  year={2026}
}

@article{bi2026skillmining,
  title={Automating Skill Acquisition through Large-Scale Mining of Open-Source Agentic Repositories: A Framework for Multi-Agent Procedural Knowledge Extraction},
  author={Bi, Shuzhen and Wu, Mengsong and Hao, Hao and Li, Keqian and Liu, Wentao and Song, Siyu and Zhao, Hongbo and Zhou, Aimin},
  journal={arXiv preprint arXiv:2603.11808},
  year={2026}
}

@article{khanda2026crystallization,
  title={Adaptive Memory Crystallization for Autonomous AI Agent Learning in Dynamic Environments},
  author={Khanda, Rajat and Chakrabarti, Mohammad Baqar Sambuddha and Changdar, Satyasaran},
  journal={arXiv preprint arXiv:2604.13085},
  year={2026}
}

@article{buyya2026agentic,
  title={Agentic Artificial Intelligence (AI): Architectures, Taxonomies, and Evaluation of Large Language Model Agents},
  author={Buyya, Rajkumar and others},
  journal={arXiv preprint arXiv:2601.12560},
  year={2026}
}

@article{plaat2025agentic,
  title={Agentic large language models, a survey},
  author={Plaat, Aske and van Duijn, Max and Van Stein, Niki and Preuss, Mike and van der Putten, Peter and Batenburg, Kees Joost},
  journal={Journal of Artificial Intelligence Research},
  volume={84},
  year={2025}
}

@article{zhang2025landscape,
  title={The landscape of agentic reinforcement learning for llms: A survey},
  author={Zhang, Guibin and Geng, Hejia and Yu, Xiaohang and Yin, Zhenfei and Zhang, Zaibin and Tan, Zelin and Zhou, Heng and Li, Zhongzhi and Xue, Xiangyuan and Li, Yijiang and others},
  journal={arXiv preprint arXiv:2509.02547},
  year={2025}
}

@article{zhao2025llm,
  title={Llm-based agentic reasoning frameworks: A survey from methods to scenarios},
  author={Zhao, Bingxi and Foo, Lin Geng and Hu, Ping and Theobalt, Christian and Rahmani, Hossein and Liu, Jun},
  journal={arXiv preprint arXiv:2508.17692},
  year={2025}
}

@inproceedings{wu2025agentic,
  title={Agentic reasoning: A streamlined framework for enhancing llm reasoning with agentic tools},
  author={Wu, Junde and Zhu, Jiayuan and Liu, Yuyuan and Xu, Min and Jin, Yueming},
  booktitle={Proceedings of the 63rd Annual Meeting of the Association for Computational Linguistics (Volume 1: Long Papers)},
  pages={28489--28503},
  year={2025}
}

@article{wei2026agentic,
  title={Agentic reasoning for large language models},
  author={Wei, Tianxin and Li, Ting-Wei and Liu, Zhining and Ning, Xuying and Yang, Ze and Zou, Jiaru and Zeng, Zhichen and Qiu, Ruizhong and Lin, Xiao and Fu, Dongqi and others},
  journal={arXiv preprint arXiv:2601.12538},
  year={2026}
}

@article{hao2026brain,
  title={Brain-Inspired Graph Multi-Agent Systems for LLM Reasoning},
  author={Hao, Guangfu and Dai, Yuming and Qin, Xianzhe and Yu, Shan},
  journal={arXiv preprint arXiv:2603.15371},
  year={2026}
}

@article{chen2026think,
  title={Think Deep, Not Just Long: Measuring LLM Reasoning Effort via Deep-Thinking Tokens},
  author={Chen, Wei-Lin and Peng, Liqian and Tan, Tian and Zhao, Chao and Chen, Blake JianHang and Lin, Ziqian and Go, Alec and Meng, Yu},
  journal={arXiv preprint arXiv:2602.13517},
  year={2026}
}

@article{yang2026tooltree,
  title={Tooltree: Efficient llm agent tool planning via dual-feedback monte carlo tree search and bidirectional pruning},
  author={Yang, Shuo and Han, Soyeon Caren and Ding, Yihao and Wang, Shuhe and Hoy, Eduard},
  journal={arXiv preprint arXiv:2603.12740},
  year={2026}
}

@article{li2026benchmark,
  title={Benchmark Test-Time Scaling of General LLM Agents},
  author={Li, Xiaochuan and Ming, Ryan and Setlur, Pranav and Paladugu, Abhijay and Tang, Andy and Kang, Hao and Shao, Shuai and Jin, Rong and Xiong, Chenyan},
  journal={arXiv preprint arXiv:2602.18998},
  year={2026}
}

@article{feng2026idrbench,
  title={IDRBench: Interactive Deep Research Benchmark},
  author={Feng, Yingchaojie and Huang, Qiang and Xie, Xiaoya and Yang, Zhaorui and Yu, Jun and Chen, Wei and Tung, Anthony KH},
  journal={arXiv preprint arXiv:2601.06676},
  year={2026}
}

@article{hu2026agentic,
  title={Agentic Tool Use in Large Language Models},
  author={Hu, Jinchao and Zhong, Meizhi and Chen, Kehai and Bai, Xuefeng and Zhang, Min},
  journal={arXiv preprint arXiv:2604.00835},
  year={2026}
}

@article{lin2026w,
  title={W\&D: Scaling Parallel Tool Calling for Efficient Deep Research Agents},
  author={Lin, Xiaoqiang and Liew, Jun Hao and Savarese, Silvio and Li, Junnan},
  journal={arXiv preprint arXiv:2602.07359},
  year={2026}
}
}



\newpage
\appendix

\section{Algorithmic Details}
\label{app:algo}

In this section, we provide the full algorithmic procedure of SAPO in
Algorithm~\ref{alg:sapo}.

\begin{algorithm}[h]
\caption{Skill-Augmented Policy Optimization (SAPO)}
\label{alg:sapo}
\begin{algorithmic}[1]
\State \textbf{Inputs:} policy $\pi_\theta$, long-term skill bank $\mathcal{B}_{\mathrm{long}}$, temporary skill bank $\mathcal{B}_{\mathrm{temp}}$, training distribution $\mathcal{D}$, rollout budget $G$, unit size $K$, temporary horizon $T_{\mathrm{prom}}$, skill-generation prompt $X_{\mathrm{skill}}$

\For{training step $t = 1,2,\dots$}
    \State Sample a batch of skill induction units $\{\mathcal{X}_i\} \sim \mathcal{D}$,
    where each unit is $\mathcal{X}_i=\{x_{i,j}\}_{j=1}^{K}$
    
    \For{each induction unit $\mathcal{X}_i$}
        \For{each query $x_{i,j} \in \mathcal{X}_i$}
            \State Retrieve skills
            $
            \mathcal{S}_{i,j}
            \leftarrow
            \mathrm{Retr}(x_{i,j}, \mathcal{B}_{\mathrm{long}} \cup \mathcal{B}_{\mathrm{temp}})
            $
            and rerank them with $\texttt{Score}_{\pi_\theta}$ defined in Eq.~\eqref{eq:policy_score}
            \State Sample base trajectories
            $
            \mathcal{Y}_{i,j}^{\mathrm{base}}
            \sim
            \pi_\theta(\cdot \mid x_{i,j}, \mathcal{S}_{i,j})
            $
        \EndFor
        
        \State Generate one candidate skill from the induction unit using Eq.~\eqref{eq:shared_skill_gen}
        
        \For{each query $x_{i,j} \in \mathcal{X}_i$}
            \State Sample skill-augmented trajectories
            $
            \mathcal{Y}_{i,j}^{\mathrm{skill}}
            \sim
            \pi_\theta(\cdot \mid x_{i,j}, \mathcal{S}_{i,j}\cup\{\hat{s}_i\})
            $
            \State Compute prompt-specific utility
            $
            u(x_{i,j},\hat{s}_i)
            $
            using Eq.~\eqref{eq:instance_utility}
        \EndFor
        
        \State Aggregate cross-prompt utility
        $
        U_{\hat{s}_i}
        =
        \frac{1}{K}
        \sum_{j=1}^{K}
        u(x_{i,j},\hat{s}_i)
        $
        using Eq.~\eqref{eq:group_utility}
        \State Store $(\hat{s}_i,U_{\hat{s}_i})$ in $\mathcal{B}_{\mathrm{temp}}$
    \EndFor

    \State Update $\pi_\theta$ for the agent task with GRPO using all collected base and skill-augmented rollouts

    \If{$t \bmod T_{\mathrm{prom}} = 0$}
        \State Update $\pi_\theta$ as the skill generator with Eq.~\eqref{eq:utility_weighted_generator}
        and distill the reduced-input skill-likelihood score with Eq.~\eqref{eq:kd}
        \State Promote high-utility and non-redundant skills from $\mathcal{B}_{\mathrm{temp}}$ to $\mathcal{B}_{\mathrm{long}}$ using Eq.~\eqref{eq:skill_promotion}
        \State Discard unpromoted temporary skills and reset $\mathcal{B}_{\mathrm{temp}}$
        \If{$|\mathcal{B}_{\mathrm{long}}|$ exceeds the bank capacity}
            \State Remove low-scoring old skills using $\texttt{Score}_{\pi_\theta}$ defined in Eq.~\eqref{eq:policy_score}
        \EndIf
    \EndIf
\EndFor
\end{algorithmic}
\end{algorithm}

\section{Prompts}
\label{app:prompt}

In this section, we provide the system prompt used for skill generation in
SAPO. The same prompt template is used for both single-query induction
($K=1$) and grouped-query induction ($K>1$). In implementation, the grouped
case does not introduce a new skill-generation interface. Instead, SAPO
formats the inputs in the same slots as the single-query case. When $K=1$,
\texttt{\{task\_description\}} contains one task and
\texttt{\{base\_trajectories\_text\}} contains its corresponding base
rollouts. When $K>1$, SAPO simply concatenates the $K$ related tasks and
their base rollouts into the same fields, for example,

\[
\texttt{\{task\_description\}}
=
\texttt{Task 1: } x_{i,1}
\ \Vert\ 
\cdots
\ \Vert\
\texttt{Task K: } x_{i,K},
\]

and

\[
\begin{aligned}
\texttt{\{base\_trajectories\_text\}}
=&\ \texttt{Trajectories for Task 1: } \mathcal{Y}^{\mathrm{base}}_{i,1}
\\
&\ \Vert\ \cdots\ \Vert\
\texttt{Trajectories for Task K: } \mathcal{Y}^{\mathrm{base}}_{i,K}.
\end{aligned}
\]

Thus, grouped-query induction only provides richer evidence to the same
prompt template. The model is still asked to return exactly one JSON object,
so the output remains a single shared skill $\hat{s}_i$ for the whole
induction unit $\mathcal{X}_i=\{x_{i,j}\}_{j=1}^{K}$.

\begin{tcolorbox}[title=System Prompt for Skill Generation Agent of SAPO, colback=gray!5!white, colframe=black!75!white, fonttitle=\bfseries]
\small
\textbf{SKILL GENERATION AGENT SYSTEM PROMPT}

You are an expert agent trainer. Based on the following task and trajectory rollouts (with rewards), generate a NEW actionable skill that would improve the agent's performance.

\vspace{0.5em}
\textbf{\#\# Task}

\{task\_description\}

\vspace{0.5em}
\{context\_skills\_text\}

\vspace{0.5em}
\textbf{\#\# Trajectories}

\{base\_trajectories\_text\}

\vspace{0.5em}
\textbf{\#\# Instructions}

Analyze the trajectories above. Identify patterns: which actions led to success vs failure? Generate exactly ONE new skill that would help the agent succeed on this type of task.

The skill must be a JSON object with these fields:
\begin{itemize}
    \item \texttt{"skill\_id"}: \texttt{"candidate"}
    \item \texttt{"title"}: 4--8 words clearly naming the strategy
    \item \texttt{"principle"}: 3--5 sentences describing WHAT to do, HOW to do it step-by-step, and WHY it works. Be specific and actionable.
    \item \texttt{"when\_to\_apply"}: 2--3 sentences specifying exact conditions that trigger this skill.
\end{itemize}

Return ONLY a valid JSON object (not an array), no other text.

\end{tcolorbox}

\section{Implementation Details}
\label{app:details}

We train \texttt{Qwen2.5-7B-Instruct} and \texttt{Qwen3-4B-Instruct} on one
node with 8 H200 GPUs. All RL experiments are implemented with the
\texttt{verl} actor-rollout-reference framework using GRPO as the advantage
estimator. The actor is optimized with AdamW using learning rate
$1\times10^{-6}$. Rollouts are
served with \texttt{vLLM} using tensor parallel size 1, and validation responses
are sampled with temperature 0.4.

\paragraph{Benchmarks and Training Setup.}
For ALFWorld, we use the GiGPO training split~\citep{fenggroup}. For WebShop,
we use the standard WebShop environment. For Search-QA, we follow
Search-R1~\citep{jin2025search}, use E5~\citep{wang2022text} as the retriever,
train on NQ and HotpotQA, and evaluate out-of-domain on TriviaQA, PopQA, 2Wiki,
MuSiQue, and Bamboogle. The Search-QA policy is initialized from
\texttt{Jianwen/Search-7B-SFT}. The main training hyperparameters are summarized
in Table~\ref{tab:implementation_hparams}.

\begin{table}[h]
\centering
\small
\begin{tabular}{lccc}
\toprule
Setting & ALFWorld & WebShop & Search-QA \\
\midrule
Train batch size & 16 & 16 & 512 \\
Validation batch size & 256 & 64 & 1,024 \\
Rollouts per prompt & 8 & 8 & 8 \\
Max prompt length & 4,096 & 6,000 & 5,000 \\
Max response length & 512 & 768 & 700 \\
Max environment steps & 50 & 15 & 4 \\
KL coefficient & 0.01 & 0.01 & 0.001 \\
Training epochs & 150 & 300 & 1 \\
Evaluation frequency & 5 & 5 & 50 \\
\bottomrule
\end{tabular}
\caption{Main training hyperparameters for SAPO.}
\label{tab:implementation_hparams}
\end{table}

\paragraph{SkillBank and Skill-Generator Settings.}
Across all environments, skills are
retrieved by embedding similarity with top-$k=3$, and SAPO maintains a dual-bank
memory with a temporary candidate bank and a long-term bank. Candidate skills
are promoted every 5 epochs after utility-based validation. We use a maximum
long-term bank size of 45 and deduplication threshold 0.8. The main SkillBank
hyperparameters are shown in Table~\ref{tab:skillbank_hparams}.

\begin{table}[h]
\centering
\small
\begin{tabular}{lccc}
\toprule
Setting & ALFWorld & WebShop & Search-QA \\
\midrule
Promotion ratio $\rho$ & 0.2 & 0.2 & 0.1 \\
Promotion interval & 5 & 5 & 5 \\
Prompt examples per skill & 4 & 4 & 16 \\
Top-$k$ retrieved skills & 3 & 3 & 3 \\
Reranking pool size & 6 & 10 & 6 \\
Max long-term bank size & 45 & 45 & 45 \\
Deduplication threshold & 0.8 & 0.8 & 0.8 \\
\bottomrule
\end{tabular}
\caption{SkillBank and auxiliary skill-generator hyperparameters.}
\label{tab:skillbank_hparams}
\end{table}

\paragraph{Hyperparameter Selection.}
Hyperparameters are selected based on validation performance and compute
constraints. We set $K=4$ queries per skill induction unit to balance
utility-estimation stability and rollout cost. We use small promotion ratios
($\rho=20\%$ for ALFWorld/WebShop and $\rho=10\%$ for Search-QA) to keep the
SkillBank compact. The deduplication threshold is set to 0.8 to filter
near-duplicate skills.

\section{Additional Experimental Results}

\subsection{Utility of Claude-Generated Skills}
\label{app:claude_utility}

To examine whether the mixed utility of generated skills is specific to
GPT-generated skills, we repeat the preliminary utility analysis in
Sec.~\ref{sec:pre} using Claude-Opus-4.6 as the skill generator. We follow the
same experimental protocol: for each prompt, we first collect base rollouts,
ask Claude-Opus-4.6 to generate a candidate skill from the observed failures,
and then collect skill-augmented rollouts with the generated skill added. The
reward gap between the two rollout groups is used as the marginal utility of the
generated skill.

\begin{figure}[t]
    \centering
    \includegraphics[width=0.8\linewidth]{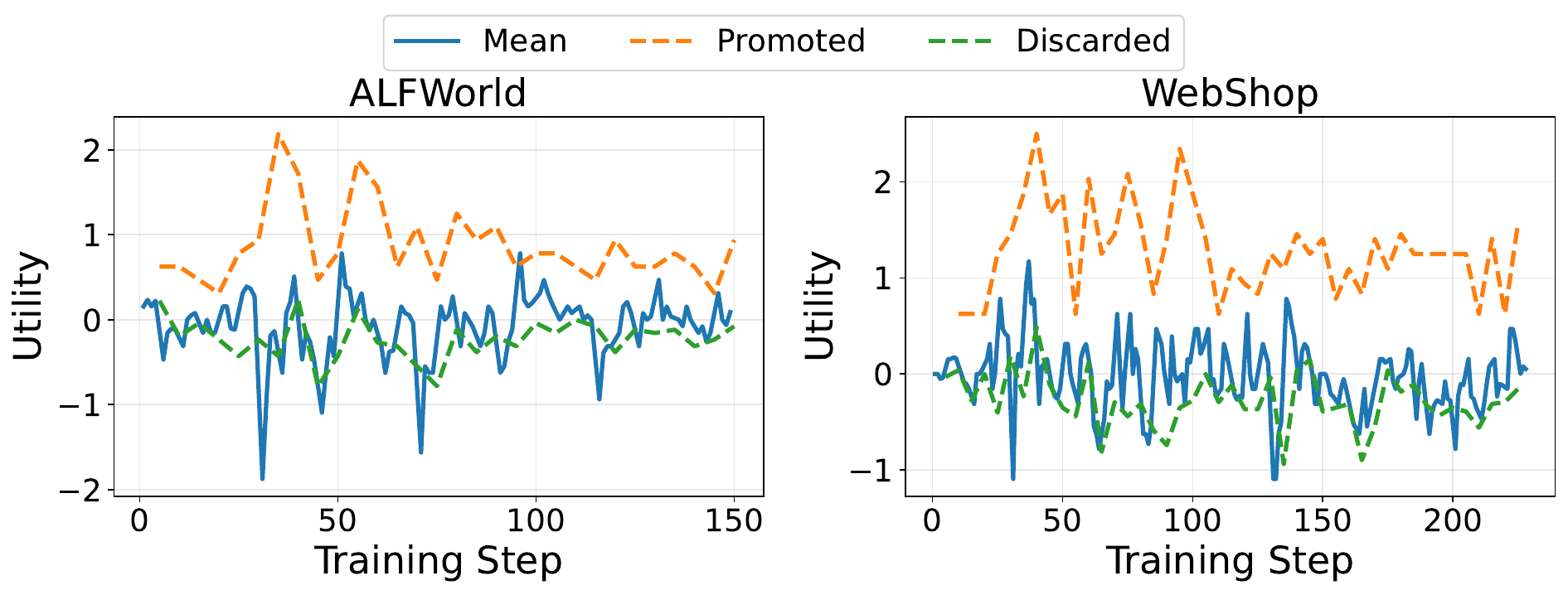}
    \caption{Utility of Claude-generated skills on ALFWorld and WebShop during training.}
    \label{fig:claude_utility}
\end{figure}

As shown in Fig.~\ref{fig:claude_utility}, Claude-generated skills also exhibit
highly mixed utility on both ALFWorld and WebShop. Although promoted skills
achieve clearly positive utility, the mean utility remains close to zero during
training, indicating that many generated skills provide limited or even negative
benefit. This result further supports our main observation that relying on
frontier LLMs does not guarantee high-quality skills, and motivates SAPO's
validate-before-store design.

\subsection{Main Results with Qwen3-4B-Instruct}
\label{app:qwen3_4b_results}

To further evaluate SAPO under a smaller backbone, we follow the experimental
setting in Sec.~\ref{sec:exp} and replace the base model with
\texttt{Qwen3-4B-Instruct-2507}. Table~\ref{tab:alfworld_qwen3_4b} reports the
results on ALFWorld. SAPO achieves the best overall performance, improving the
All score from 72.7 to 82.0 over the strongest baseline. These results show that
SAPO's gains are not limited to the 7B backbone used in the main experiments and
remain effective with a smaller open-weight policy.
\begin{table}[h]
\centering
\small
\caption{ALFWorld performance using \texttt{Qwen3-4B-Instruct-2507} as the base model.}
\begin{tabular}{lccccccc}
\toprule
\textbf{Method} & \textbf{Pick} & \textbf{Clean} & \textbf{Cool} & \textbf{Look} & \textbf{Heat} & \textbf{Pick2} & \textbf{All} \\
\midrule
Origin & 50.0 & 9.5 & 0.0 & 2.1 & 11.1 & 4.8 & 17.2 \\
GRPO & 73.5 & 46.6 & 48.0 & 61.1 & 62.5 & 20.0 & 53.9 \\
SkillRL (O3) & 90.0 & 92.3 & 52.0 & 63.6 & 42.9 & 40.9 & 67.2 \\
D2Skill (Gemini-3-Flash) & 88.6 & 75.0 & 54.2 & 66.7 & 60.0 & 52.6 & 69.6 \\
D2Skill (O3) & 89.4 & 72.4 & 66.7 & 54.5 & 60.0 & 50.0 & 72.7 \\
SAPO & 86.9 & 93.1 & 82.2 & 77.7 & 89.6 & 67.3 & 82.0 \\
\bottomrule
\end{tabular}
\label{tab:alfworld_qwen3_4b}
\end{table}


\subsection{Training Dynamics on ALFWorld Subtasks}
\label{app:alfworld_subtask_dynamics}

We further analyze the training dynamics of SAPO and SkillRL on individual
ALFWorld subtasks. Following the experimental setting in Sec.~\ref{sec:exp}, we
use \texttt{Qwen2.5-7B-Instruct} as the base model and report validation
performance throughout training.

\begin{figure}
    \centering
    \includegraphics[width=0.95\linewidth]{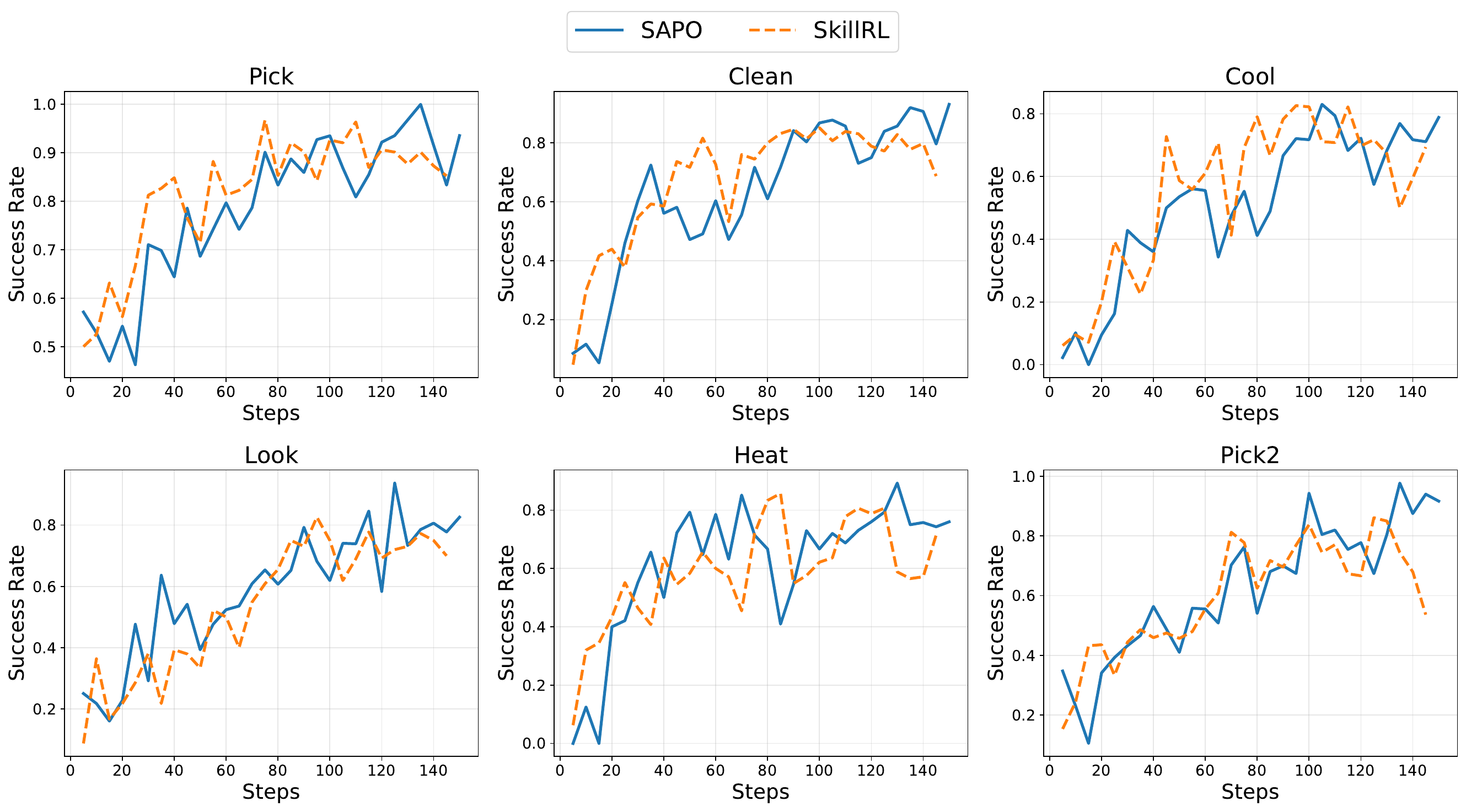}
    \caption{Training dynamics of SAPO and SkillRL on ALFWorld subtasks using
    \texttt{Qwen2.5-7B-Instruct} as the base model.}
    \label{fig:alfworld_subscore}
\end{figure}

As shown in Fig.~\ref{fig:alfworld_subscore}, SAPO exhibits steady performance
improvement across ALFWorld subtasks and shows smaller fluctuations in the later
stage of training. In contrast, SkillRL often declines after reaching its peak
performance on several subtasks. This suggests that directly evolving the skill
bank without explicit skill curation can introduce low-quality or outdated
skills, which may mislead the agent and destabilize later-stage learning.

\subsection{Hyperparameter Analysis}
\label{app:hyperparameter}

We study the effect of $K$, the number of queries used to induce each candidate
skill. Table~\ref{tab:hyperparameter} reports results with different values of
$K$ and includes SkillRL as the baseline. SAPO consistently outperforms SkillRL
across all tested settings, showing that its gains are not sensitive to a
specific choice of $K$. Using multiple related queries generally improves over
single-query induction, suggesting that grouped induction provides richer
evidence for generating reusable skills. We use $K=4$ as the default setting,
which achieves the best WebShop score and strong overall performance across both
benchmarks.

\begin{table}[h]
\centering
\small
\setlength{\tabcolsep}{6pt}
\caption{Hyperparameter analysis of $K$, the number of queries used to induce
each candidate skill.}
\begin{tabular}{lccc}
\toprule
\multirow{2}{*}{Method} & ALFWorld & \multicolumn{2}{c}{WebShop} \\
\cmidrule(lr){2-2} \cmidrule(lr){3-4}
& All. & Score & Succ. \\
\midrule
SkillRL (Baseline) & 89.9 & 85.2 & 72.7 \\
SAPO ($K=1$) & 91.4 & 91.2 & 76.5 \\
SAPO ($K=8$) & 93.3 & 88.8 & 79.6 \\
\midrule
SAPO ($K=4$) & 92.2 & 90.5 & 78.1 \\
\bottomrule
\end{tabular}
\label{tab:hyperparameter}
\end{table}

\section{API Cost of Baseline Skill Generation}
\label{app:api_cost}

SkillRL-style skill evolution relies on proprietary frontier LLM calls to
generate new skills during training. In our ALFWorld runs, using GPT-5.4 for
skill generation costs roughly \$30 for a single run, even before accounting for
environment rollout, retrieval, or policy-optimization cost. While this cost may
appear modest at small scale, it becomes non-trivial for individual researchers
when repeated across benchmarks, seeds, ablations, and hyperparameter settings.
In contrast, SAPO trains the policy itself as the skill generator and derives
skill-utility signals from the same rollouts used for agent learning, thereby
avoiding repeated proprietary LLM calls during online RL.


\newpage

\end{document}